\documentclass[11pt]{article}
\usepackage{jmlr}

\usepackage{amsmath,amsfonts,amssymb,bm}

\def\Figref#1{Figure~\ref{#1}}

\def\Secref#1{Section~\ref{#1}}

\def\eqref#1{equation~\ref{#1}}
\def\Eqref#1{Equation~\ref{#1}}

\def\Algref#1{Algorithm~\ref{#1}}

\def\1{\bm{1}}

\DeclareMathAlphabet{\mathsfit}{\encodingdefault}{\sfdefault}{m}{sl}
\SetMathAlphabet{\mathsfit}{bold}{\encodingdefault}{\sfdefault}{bx}{n}

\newcommand{\softmax}{\mathrm{softmax}}

\newcommand{\algcomment}[1]{{\footnotesize \fontfamily{cmtt}\selectfont // #1}}

\usepackage{graphicx}
\usepackage{subcaption}
\usepackage{booktabs}
\usepackage[most]{tcolorbox}
\usepackage{nicefrac}
\usepackage{algorithm}
\usepackage{algorithmic}
\usepackage{wrapfig}
\usepackage[export]{adjustbox}
\usepackage{pdflscape}

\begin{document}

\title{Plug \& Play Directed Evolution of Proteins with Gradient-based Discrete MCMC}

\author{\name Patrick Emami \email pemami@nrel.gov \\
      \addr National Renewable Energy Lab
      \AND
        \name Aidan Perreault \email aperr@stanford.edu\\
      \addr Stanford University
        \AND
      \name Jeffrey Law \email jeffrey.law@nrel.gov\\
      \addr National Renewable Energy Lab
      \AND
      \name David Biagioni \email dave@maplewelleng.com \\
      \addr Maplewell Energy
      \AND
      \name Peter C. St. John \email me@pcstj.com\\
      \addr National Renewable Energy Lab
      }

\editor{}

\maketitle

\begin{abstract}
A long-standing goal of machine-learning-based protein engineering is to accelerate the discovery of novel mutations that improve the function of a known protein.
We introduce a sampling framework for evolving proteins \emph{in silico} that supports mixing and matching a variety of unsupervised models, such as protein language models, and supervised models that predict protein function from sequence.
By composing these models, we aim to improve our ability to evaluate unseen mutations and constrain search to regions of sequence space likely to contain functional proteins. 
Our framework achieves this without any model fine-tuning or re-training by constructing a product of experts distribution directly in discrete protein space. 
Instead of resorting to brute force search or random sampling, which is typical of classic directed evolution, we introduce a fast MCMC sampler that uses gradients to propose promising mutations. 
We conduct \emph{in silico} directed evolution experiments on wide fitness landscapes and across a range of different pre-trained unsupervised models, including a 650M parameter protein language model.
Our results demonstrate an ability to efficiently discover variants with high evolutionary likelihood as well as estimated activity multiple mutations away from a wild type protein, suggesting our sampler provides a practical and effective new paradigm for machine-learning-based protein engineering.
\end{abstract}	

\section{Introduction}
\label{sec:intro}
\begin{figure}[t]
    \centering
    \begin{subfigure}{\textwidth}
        \centering
        \includegraphics[scale=0.6]{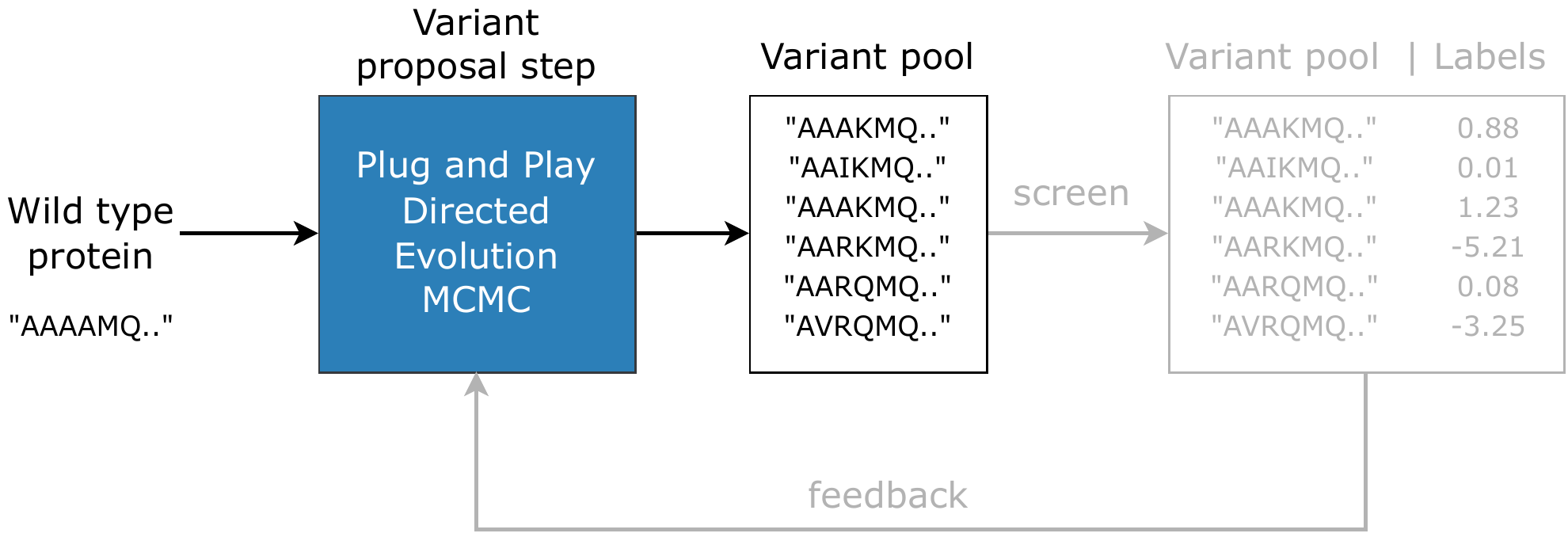}
        \caption{\label{fig:de}}
        \vspace{3mm}
    \end{subfigure}
    \begin{subfigure}{\textwidth}
        \centering
        \includegraphics[scale=0.7]{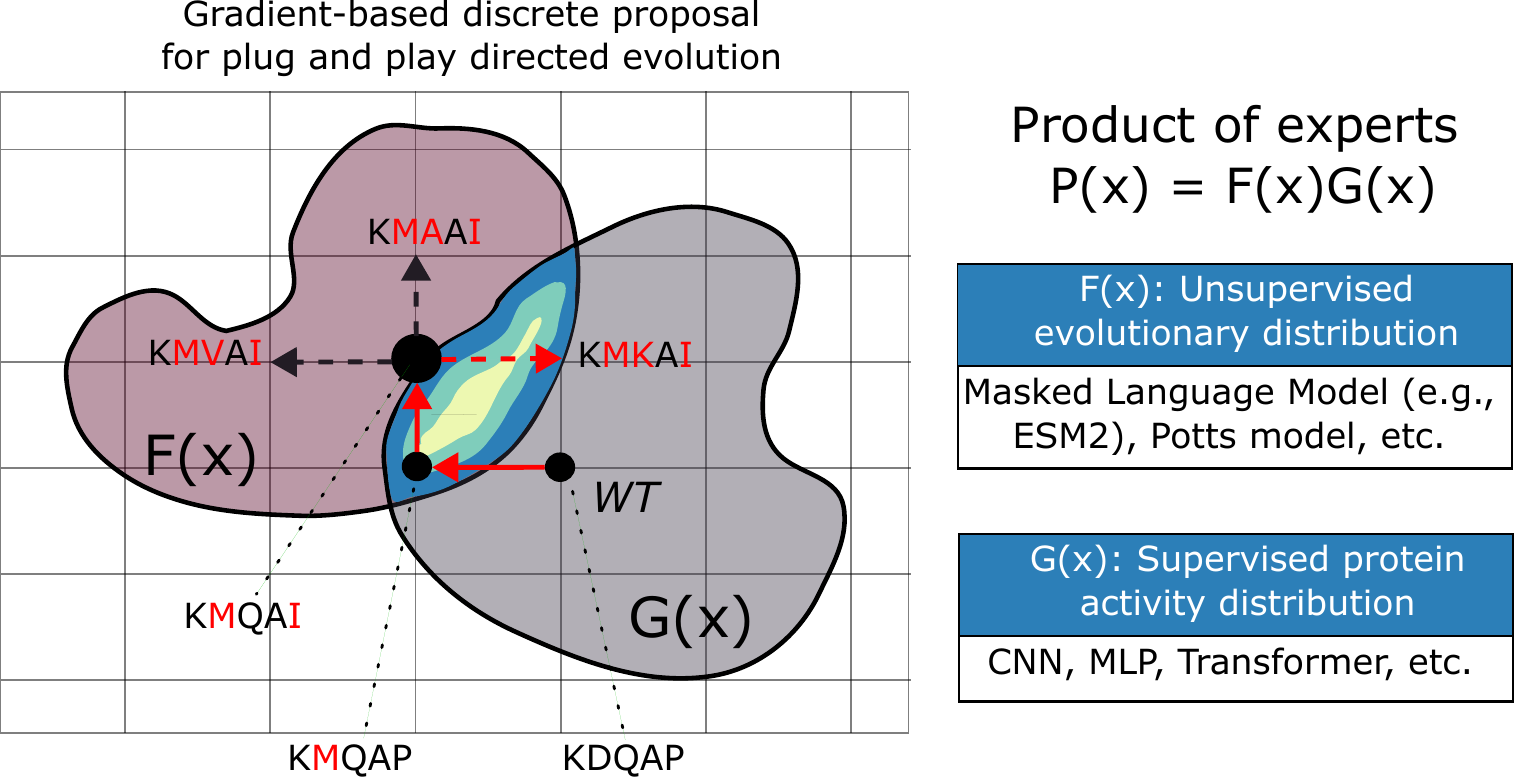}
        \caption{\label{fig:pe}}
    \end{subfigure}
    \caption{a) Illustration of the machine-learning-based directed evolution pipeline. This paper proposes a plug and play sampler for proposing variants of improved fitness near a wild-type (WT) protein for each design round.
    b) We flexibly compose unsupervised and supervised protein sequence models  \emph{as a product of experts} distribution and derive a fast gradient-based discrete MCMC sampler to efficiently sample from this distribution.
    A product of experts places its probability mass at the \emph{intersection} of the expert distributions. 
    The key idea of gradient-based discrete MCMC is to bias the mutation proposal distribution towards promising mutations within a local neighborhood around the current variant.
    Then, to avoid enumerating all mutations in this neighborhood, the proposal is approximated with a first-order Taylor series.
    An example MCMC trajectory of single point mutations starting from the WT is shown.}
    \label{fig:1}
\end{figure}
Engineering proteins to improve their productivity or catalyze new reactions requires scientists to navigate the complex landscape mapping a protein's amino acid sequence to its structure and function \citep{li2020protein}.
\emph{Directed evolution} is a classic approach inspired by natural evolution where random mutations to a protein's sequence are screened in a wet lab until higher-performing variants are found, at which point the process repeats starting from these variants~\citep{Kuchner_1997}.
However, this becomes impractical when proteins cannot be assessed in a high-throughput fashion.
The simplest approach, brute force search, is limited in practice to variants with one or two mutations.
For a protein with 400 amino acids, there are $\sim 10^{19}$ ways to make five single substitutions assuming the standard vocabulary of 20 amino acids.
It is also remarkably difficult to find proteins with improved function.
Most of protein space is non-functional and beneficial mutations are rare~\citep{arnold1998design}.

As supervised machine learning methods for protein function prediction from primary sequence improve~\citep{dallago2021flip,hsu2022learning}, \emph{machine-learning-based directed evolution} has emerged, which offers a way to judiciously propose candidates for screening that ultimately reduces time spent in the wet lab~\citep{Yang2019,wu2021protein,Biswas2021}.
Our work is concerned with improving the variant proposal step by increasing the chance of discovering a mutation that improves a target property of a known ``wild type'' (WT) protein (\Figref{fig:de}).
We argue that the promising performance of recent unsupervised sequence models at mutation effect prediction~\citep{Meier2021,hsu2022learning,weinstein2022nonidentifiability} has motivated a reconsideration of simple black-box optimization algorithms for searching sequence space.
Black-box algorithms are appealing due to their simplicity, flexibility, compatibility with discrete spaces, and use of interpretable mutation operators.
They offer a way to mix and match unsupervised and supervised models for proposing variants without requiring any model fine-tuning or re-training, i.e., in a ``plug and play'' manner.
Combining both types of models is generally advantageous.
While unsupervised models learn information that can steer search away from adversarial inputs which fool supervised models due to overestimation errors~\citep{Szegedy2014}, supervised models learn specific information about beneficial mutations gleaned from assay-labeled data.
However, black-box algorithms tend to be extremely inefficient at searching for variants with improved fitness.
A practical plug and play framework for searching protein sequence space has remained elusive due to the difficulty of fast search in discrete, high-dimensional spaces.

This paper fills that gap by introducing Plug and Play Directed Evolution (PPDE), a practical plug and play sampling framework for the efficient discovery of functional variants directly in discrete protein space.
PPDE flexibly combines unsupervised and supervised models with a \emph{product of experts} distribution~\citep{hinton2002training} $p(x) = \prod_{i} p_i(x)$.
Combining both types of protein models in this manner encourages variants to have both high sequence likelihood and high predicted function (e.g., activity).
To sample efficiently from the high-dimensional, discrete, and unnormalized distribution $p(x)$, we derive a fast Markov chain Monte Carlo (MCMC) sampler that uses \emph{gradients} of $p(x)$ to propose mutations.
See \Figref{fig:pe} for a visual depiction the sampler.
PPDE nearly maintains all of the flexibility of black-box algorithms---it only has to additionally assume that $p(x)$ is differentiable at each discrete point of protein space.
We theoretically characterize the efficiency of this sampler and empirically show that our framework works with a variety of pre-trained models without using continuous relaxations and without model retraining or fine-tuning.
Although our formulation can be applied to a broad class of biological sequence design tasks, we focus on proteins due to the current widespread availability of pre-trained models.

A summary of our contributions:
\begin{itemize}
    \item We introduce a plug and play sampling framework that enables mixing and matching various unsupervised and supervised protein models without model re-training or fine-tuning.
    \item We demonstrate a novel application of gradient-based discrete MCMC for conducting search with a product of experts distribution. 
    \item We conduct rigorous \emph{in silico} directed evolution experiments on three target proteins with partially characterized wide ($\geq$ 2 mutation) fitness landscapes. Our results provide strong evidence that our sampler offers a practical and effective approach for ML-based directed evolution beyond brute force and random search.
\end{itemize}

\section{Related Work}
\label{sec:relwork}

\textbf{Black-box approaches to directed evolution:} A popular paradigm for ML-based directed evolution has been to learn a surrogate for the sequence-to-function landscape from a small labeled dataset and then to use a black-box optimizer or sampler such as an evolutionary algorithm or simulated annealing to search for mutants~\citep{hansen1996adapting,Angermueller,Sinai2020,Biswas2021}. 
Black-box approaches are plug and play in that they can easily combine unsupervised and supervised protein models without any model fine-tuning or re-training.
Navigating the high-dimensional discrete space is approached by either using a random walk, which is highly inefficient, or by using gradient ascent via a continuous relaxation of the space, which can bias the search process towards poor local optima.
These are the main baselines for our method and will be described in more detail in Section~\ref{sec:experiments}.
Reinforcement learning (RL) has also been used as a black-box optimization framework for training autoregressive models to generate high-fitness variants~\citep{angermueller2019model}.
However, this approach has less flexibility than plug and play methods, in that it does not support easy mixing of unsupervised pre-trained models with the learned RL policy.

\textbf{Protein design by generative modeling:} An alternative to directed evolution is to cast protein design as inference, i.e., fitting a conditional generative model to assay-labeled proteins showing high activity~\citep{brookes2018design,Brookes2021,Fannjiang,Gupta2018,zhang2022unifying,jain2022biological}. 
Since high performing variants are rare, this task amounts to density estimation of a rare event which is a difficult statistical inference problem in its own right.
Another popular alternative is latent space optimization, in which gradient ascent is performed on a surrogate function trained directly in the latent space of a generative model.
These generative models use regularization to shape their latent space favorably so that optimization remains constrained to the manifold of viable proteins~\citep{Killoran2017,GomezBombarelli2018,Kumar2019,Linder2020,chan2021deep,Trabucco2021,Castro2022}.
The Deep Manifold Sampler (DMS)~\citep{Gligorijevic2021} more explicitly constrains optimization by alternating between steps of gradient ascent and re-projection of a noised version of the sequence onto the latent space of a custom denoising autoencoder (DAE).
Our plug and play approach works with a variety of unsupervised models and can combine multiple such models if desired.

At the expense of trust in candidate proteins, unconditional generative models can also be used to hallucinate proteins from distant, unexplored regions of protein space.
For example, training a generative model on sequences from a target protein family has been used to generate functional variants~\citep{Costello2019,HawkinsHooker2021,Shin2021}.
Unconditional sampling from protein language models trained on unaligned~\citep{Madani2020,rives2021biological,Hesslow2022,lin2022language,Ferruz2022,Nijkamp2022,Yang2022} and aligned~\citep{rao2021msa,Notin2022} sequences has recently been explored.
While this is a promising new direction for protein engineering, our focus is on directed evolution from known proteins.

\textbf{Plug and play text generation:} Controlled generation of text has similarly been approached by combining pre-trained unsupervised models (e.g., large language models) and supervised models (e.g., sentiment classifiers) as a discrete product of experts~\citep{holtzman2018learning,Dathathri2020,Qin2022}.
Unlike controlled text generation, where the goal is to guide a sampling process to transform a sentence by altering entire words and phrases, \emph{directed evolution seeks to accumulate a few precise changes to an initial sequence by modifying a small number of amino acids} (i.e., individual characters).
Moreover, plug and play approaches for controlled text generation intentionally avoid solving a combinatorial search problem by instead restricting the class of suitable unsupervised experts to left-to-right autoregressive language models. 
This enables the use of sequential sampling algorithms such as left-to-right beam search decoding.
Our use of gradient-based discrete MCMC for plug and play directed evolution allows us to efficiently search the combinatorial space of mutations near the wild type protein and to use a more general class of unsupervised experts that includes \emph{orderless} protein sequence models which capture epistatic relationships between any pair of amino acids.
The Metropolis-adjusted Langevin Algorithm (MALA) used for \emph{plug and play image generation}~\citep{Nguyen2017} is a simple gradient-based MCMC sampler for continuous spaces, and is closely related to locally-balanced discrete MCMC~\citep{zhang2022langevin,Sun2022}.
Therefore, we use MALA with a hand-crafted continuous relaxation as a baseline for plug and play sampling from discrete product of experts in our experiments.

\section{Background}
\label{sec:bg}

Before presenting our sampler, we define our search problem, briefly review Metropolis-Hastings (MH) MCMC, and introduce gradient-based discrete MCMC, a class of MH samplers able to efficiently explore high-dimensional discrete distributions.

\textbf{Problem definition: }The variant proposal step of directed evolution involves searching for mutations that improve one or more target properties of a given wild type (WT) protein.
This search problem is formally defined over discrete protein sequences $x := \{x_0, \dots, x_{L-1}\}$, $x \in X$, of length $L$ with each $x_i$ taking on a value in a vocab of size $V$ (typically $V = 20$ for the 20 standard amino acids).
We assume that each $x_i$ is one-hot encoded.
The search is initialized at the WT protein $x^{WT}$ and terminates when a predefined condition is met (e.g., a maximum allowable number of search steps is reached).

\textbf{Metropolis Hastings: }The MH algorithm~\citep{metropolis1953equation,hastings1970monte} defines the following algorithm for drawing samples from a distribution over protein variants $p(x)$.
Let $f(x)$ be the unnormalized log probability of $x$ such that $\log p(x) = f(x) - \log Z$ where $Z = \sum_{x \in X} \exp(f(x))$ is a normalizing constant. 
Given the current state $x$, draw a candidate next state $x'$ from proposal distribution $q(x' \mid x)$. Accept the proposed state with probability 
\begin{equation*}
\min \biggl\{1, \exp(f(x') - f(x))\frac{q(x|x')}{q(x'|x)} \biggr\},    
\end{equation*} otherwise reject the transition and stay at $x$. 
This probabilistic acceptance/rejection criterion is desirable since it provides MH samplers with theoretical convergence guarantees and does not contain any hyperparameters, simplifying the implementation of MH in practice. 

\textbf{Gradient-based discrete MCMC:} Uninformed MH proposals such as the uniform distribution are often inefficient for sampling from high-dimensional discrete distributions since candidate states $x'$ are proposed ``blindly''.
In general, the efficiency of MH is highly dependent on the choice of proposal distribution.
MH with \emph{locally-balanced informed proposals}~\citep{zanella2020informed} use distributions of the form 
\begin{equation}
    \label{eq:zanella}
    q(x'|x) \propto \exp(f(x') - f(x))^{\frac{1}{2}}\textbf{1}(x' \in \mathcal{N}(x)),
\end{equation}
where $\textbf{1}(x' \in \mathcal{N}(x))$ is an indicator function.
The key idea is to bias the proposal distribution towards local state transitions within a neighborhood $\mathcal{N}(x)$ that incur an increase in likelihood. 
We take the square root of the exponential in this proposal as the choice of local balancing function  $w(t) = t w(1/t),  \forall t>0$. This function ``balances'' the acceptance and rejection probabilities in the local neighborhood $\mathcal{N}(x)$ to achieve a high acceptance rate. 
The square root $w(t) = \sqrt{t}$ was empirically validated as a good default option in~\citet{zanella2020informed}.
Since enumerating all local moves in $\mathcal{N}(x)$ in a discrete high-dimensional spaces is infeasible, an efficient alternative is available for functions $f(x)$ whose gradient can be evaluated at the discrete state $x$~\citep{Grathwohl2021}.
A \emph{gradient-based} locally-balanced informed proposal is the first-order Taylor-series approximation of \Eqref{eq:zanella} around $x$:
\begin{equation}
\label{eq:gwg}
\Tilde{q}(x'|x) \propto \exp \biggl( \frac{1}{2} \nabla_x f(x)^{T}(x' - x) \biggr) \textbf{1}(x' \in \mathcal{N}(x)).
\end{equation}
When $\mathcal{N}(x)$ is the 1-Hamming ball, this proposal amounts to a tempered softmax over single changes to one dimension of $x$.
One forward pass and one backwards pass is required to compute the forward $\Tilde{q}(x' | x)$ and reverse $\Tilde{q}(x |x')$ approximate proposals.

A concern with this sampler is that it is susceptible to getting trapped in poor local optima since it cannot perform large jumps across the search space at each step of MCMC~\citep{sun2022path}. 
One way to address this is to increase the Hamming window size, $U$, to $U > 1$.
While this enables the sampler to propose changes to multiple dimensions of $x$ simultaneously, it also incurs a large increase in computation since $\mathcal{N}(x)$ now contains $O( (LV)^U )$ states to evaluate.
Alternatively, a \emph{path proposal} sequentially samples $R \sim \text{Unif}(1,U)$ single changes to apply to the current state $x$~\citep{sun2022path}.
For example, if $x$ is the sequence ``AAA'', then a sampled path of length 3 could look like ``AAA'' $\rightarrow$ ``AAB'' $\rightarrow$ ``AAC'' $\rightarrow$ ``ABC''.

The approximate path proposal distribution starting at $x^0 := x$ is
\begin{align}
\label{eq:path}
\Tilde{q}_{R}(x'|x) &= \prod_{r=1}^{R} \Tilde{q}\bigl(x^r |x^{r-1} \bigr) \nonumber \\
&\propto \prod_{r=1}^R \exp \Biggl( \frac{1}{2} \nabla_{x^0} f( x^0 )^T( x^r - x^{r-1}) \Biggr) \textbf{1}(x^r \in \mathcal{N}(x^{r-1})).
\end{align}
The $r$\textsuperscript{th} path state $x^r$ is sampled from $\Tilde{q}\bigl(x^r|x^{r-1} \bigr)$.
The sampler decides whether to accept or reject the terminal path state $x' := x^{R}$ (e.g., ``ABC''), i.e., the accumulation of all $R$ changes applied to $x$.
The reverse proposal $\Tilde{q}_R(x|x')$ is computed using the gradient at the terminal state $\nabla_{x'} f(x')$ and the reversed sequence of states $(x^{R}, x^{R-1},\dots,x^0)$.
The number of forward and backwards passes required to compute the forward and reverse path proposals is still two.
In the next section, we will introduce and characterize a gradient-based path proposal for product of experts distributions.

\section{Plug and Play Directed Evolution}
\label{sec:p3o}
Our approach for proposing mutations that improve a target property of the WT protein is to sample from a distribution constructed by combining unsupervised and supervised sequence models without using any continuous relaxations, model re-training, or fine-tuning.
This distribution places its probability mass on proteins that have both high \emph{evolutionary density} $f(x)$ (i.e., high likelihood of being a naturally occurring protein) and high predicted function $g(x)$ (e.g., activity or stability) by taking the product of multiple pre-trained ``expert'' distributions:
\begin{equation}
    \label{eq:PoE}
    \log p(x) = \sum_i f_i(x) + \lambda \sum_j g_j(x) - \log Z.
\end{equation}
Each $f_i(x)$ is an unsupervised model, each $g_j(x)$ is a supervised model, and 
\begin{equation*}
Z = \sum_{x \in X} \exp( \sum_i f_i(x) + \lambda \sum_j g_j(x) )    
\end{equation*} 
is the unknown normalizing constant.
Typically, $f_i(x)$ has been trained to do density estimation on unlabeled yet aligned sequences (e.g., a collection of evolutionarily related sequences provided by a multiple sequence alignment (MSA)) or unaligned sequences.
While in this work we assume the $g_j(x)$ are an ensemble of nonlinear models trained to regress activity from a labeled dataset of mutants, this formulation can easily be extended to the multi-objective case where, for e.g., $g_1$ predicts activity and $g_2$ predicts stability. 

The unsupervised experts $p_{f}(x) \propto \prod_i \exp{ (f_i(x)) }$ act as a soft constraint that keeps the sampler near regions of high evolutionary density and away from, e.g., adversarial local optima of the supervised models.
Examples of unsupervised models that provide an evolutionary density score include the EVmutation Potts model~\citep{Hopf2017}, the ESM protein masked language models~\citep{rives2021biological,lin2022language}, denoising autoencoders (DAEs), energy based models (EBMs), autoregressive protein language models (e.g., ProGen2~\citep{Nijkamp2022}), and normalizing flows (non-exhaustive list).
The supervised expert $p_g(x)$ acts as a soft constraint that guides sampling towards proteins that have high activity, where $p_{g}(x) \propto  \prod_j \exp{ ( \lambda\hspace{.5mm} g_j(x)) }$ assigns high probability to sequences with high activity. 
The hyperparameter $\lambda \geq 0$ allows us to balance the contribution of the unsupervised and supervised experts; for example, we can emphasize the ``realism'' of the protein (the evolutionary density) by setting $\lambda$ to 0.
However, recent evidence indicates that evolutionary density scores also positively correlate with protein fitness in a ``zero-shot'' manner (i.e., without using any labels)~\citep{Meier2021,hsu2022learning,weinstein2022nonidentifiability}.

\subsection{Product of experts gradient-based discrete MCMC} 
Sampling from the product of experts (\Eqref{eq:PoE}) is difficult since the normalization constant $Z$ is assumed unknown and is intractable to compute in practice.
Although we have a good initialization for an MCMC sampler---$x^{WT}$, the wild type protein of verified viability---traditional MCMC based on random walk exploration is too inefficient to discover variants with high predicted activity in reasonable time.

Our solution is to use fast gradient-based discrete MCMC.
We need only additionally assume that each expert is a continuous function that is differentiable at each discrete $x \in X$ (e.g., as is the case when the experts are neural networks).
In detail, assume we have $M$ continuously differentiable unsupervised experts and $N$ continuously differentiable supervised experts.
During each step of MCMC, we use the gradients of the $M + N$ experts to approximate the change in product of experts likelihood due to making $R$ point mutations to the current protein variant.
This allows us to bias the proposal distribution towards the most promising $R$ mutations.
The gradient-based Taylor approximation of our MCMC proposal for $\log p(x)$ with path length $R \sim \text{Unif}(1,U)$ is $\Tilde{q}_{R}(x'|x) = \prod_{r=1}^{R} \Tilde{q}\bigl(x^r|x^{r-1} \bigr)$, where
\begin{align}
&\phantom{\propto }\Tilde{q}\bigl(x^r|x^{r-1} \bigr) \\
&\propto \exp \Biggl( \frac{1}{2} \sum_{i=1}^M \nabla_{x^0} f_i( x^0 )^T( x^r - x^{r-1}) + \frac{\lambda}{2} \sum_{j=1}^N \nabla_{x^0} g_j(x^0)^T (x^r - x^{r-1}) \Biggr) \textbf{1}(x^r \in \mathcal{N}(x^{r-1})).\nonumber
\end{align}
We use this path proposal to sample $R$ single amino acid substitutions, which we apply to the current variant $x^0 := x$ at each step of MCMC.
The terminal state of the path $x^{R} := x'$ is the variant that results from the accumulation of the $R$ substitutions.
To avoid computing extra forward and backwards passes through the product of experts for the intermediate path proposals $\Tilde{q}(x^{r} | x^{r-1}) $, following~\citet{sun2022path} we re-use the gradient taken with respect to the path origin $x^0$ instead of recomputing gradients at intermediate states.
The same is done for the reverse path proposals $\Tilde{q}(x^{r-1} | x^r)$ with respect to the terminal state $x^{R}$.
In the next section, we characterize the efficiency of our sampler to understand how composing Taylor approximations for multiple experts affects the theoretical convergence rate.

\begin{algorithm}[t]
\begin{algorithmic}
\INPUT{one-hot encoded wild-type protein $x^{WT}$, \textcolor{black}{unsupervised experts $f_i$}, \textcolor{black}{supervised experts $g_j$}, scale $\lambda$, max path length $U$}
\OUTPUT{evolved protein $x^*$}
\WHILE{\emph{still searching}}
\STATE \textbf{define} $x := x^0$, $x' := x^{R}$, $\pi(x) := \sum_i f_i(x) + \lambda \sum_j g_j(x)$
\STATE \algcomment{compute the forward path proposal distribution}
\STATE sample path length $R \sim \text{Unif}(1,U)$
\FOR{$r\in\{1,\ldots,{R}\}$}
\STATE $\Tilde{h}(x^{r-1}) = \sum_i  \nabla_{x} f_i( x )^T( x^r - x^{r-1}) +  \lambda \sum_j  \nabla_{x} g_j( x )^T( x^r - x^{r-1})$ 
\STATE $\Tilde{q}(x^r | x^{r-1}) = \mathrm{categorical} \biggl( \softmax{ \biggl( \frac{ \Tilde{h}(x^{r-1})}{2} \biggr)} \biggr)$
\STATE \algcomment{sample a single amino acid substitution and apply it to $x^{r-1}$}
\STATE $x^r \sim \Tilde{q}(x^r | x^{r-1})$  
\ENDFOR
\FOR{$r\in\{{R},\ldots,1\}$}
\STATE $\Tilde{h}(x^r) = \nabla_{x'} \sum_i f_i( x' )^T( x^{r-1} - x^r) +  \lambda \nabla_{x'}\sum_j  g_j( x' )^T( x^{r-1} - x^{r})$ 
\STATE $\Tilde{q}(x^{r-1}|x^r) = \mathrm{categorical} \biggl( \softmax{ \biggl( \frac{ \Tilde{h}(x^r) }{2} \biggr)} \biggr)$
\ENDFOR
\STATE \algcomment{accept $x'$ with probability}
$$
\min \biggl\{1, \exp(\pi(x') - \pi(x))\frac{\prod_{r={R}}^{1} \Tilde{q}(x^{r-1} | x^r)}{ \prod_{r=1}^{R} \Tilde{q}(x^{r} | x^{r-1})} \biggr\}
$$
\ENDWHILE
\end{algorithmic}

\caption{Plug and Play Directed Evolution (PPDE)}
\label{alg:pppo}
\end{algorithm}
\Algref{alg:pppo} shows pseudo-code for our fast MCMC sampler for plug and play directed evolution of proteins. 
The sampler follows the basic structure of MH MCMC.
At each sampler step, we first compute the forward path proposal distribution $\Tilde{q}_R(x'|x)$ which we use to sample the proposed protein $x'$.
Then, we compute the reverse path proposal distribution $\Tilde{q}_R(x|x')$.
We use these distributions to compute an acceptance criterion for determining whether to accept or reject $x'$, after which the process repeats until termination (e.g., a predetermined number of MCMC steps is reached).

\subsection{Sampler Analysis}
\label{sec:analysis}

Since we are using a gradient-based approximation of the product of experts proposal distribution, a natural question is whether this approximation reduces the sample efficiency of our sampler. 
It turns out that the choice of unsupervised and supervised experts plays a key role in determining the theoretical sampler efficiency.
In particular, the following corollary to Theorem 3 from~\citet{sun2022path} relates the smoothness of each expert's gradient to our sampler's ability to efficiently explore protein space.

\begin{corollary}
Assume $\lambda = 1$, each expert $h_i$ is differentiable, $\nabla_x h_i(x)$ is $K_i$-Lipschitz, the max path length is $U$, and a 1-Hamming ball neighborhood $\mathcal{N}(x)$. Let $Q_R(x, x')$ and $\Tilde{Q}_R(x,x')$ be the Markov transition kernels induced by our sampler with the product of experts proposal $q_R(x'|x)$ and with its approximation $\Tilde{q}_R(x'|x)$, respectively.
These transition kernels are related by
\begin{equation}
    \label{eq:bound}
    \Tilde{Q}_R(x,x') \geq \Biggl( \prod_{i=1}^M e^{- K_i \frac{U(U+1)}{2} } \Biggr) Q_R(x,x').
\end{equation}
\end{corollary}
See Appendix \ref{sec:app:proof} for the proof. 

\textbf{Remark: }This result bounds the efficiency of the sampler by a \emph{product of exponential functions} of each expert's gradient's Lipschitz constant $K_i$.
This means that if just one expert has a gradient with a large Lipschitz constant (e.g., the unsupervised experts are highly nonlinear protein language models), it is possible that this expert greatly reduces the overall efficiency of the sampler.
For supervised experts (possibly an ensemble), which are typically shallow neural networks whose gradients have small Lipschitz constants, we can expect that the first-order approximation will not greatly reduce the sampler's efficiency.
However, \Eqref{eq:bound} is a fairly loose bound.
We will empirically compare how the sampler fares with different types of experts in practice (see \Secref{sec:swap_gen}).

\section{Experiments}
\label{sec:experiments}
In this section, we present the results of multiple synthetic experiments.
First, we validate that our proposed sampler is able to discover a diverse set of good optima for product of experts distributions in high-dimensional discrete spaces using a toy MNIST-based task.
Then, with a realistic \emph{in silico} directed evolution experimental setup, we characterize advantages of combining unsupervised and supervised models and compare our sampler with appropriate baseline plug and play algorithms.
The baselines are:
\begin{itemize}
    \item \textbf{Simulated annealing:} This is a simple random-mutation-based MCMC-style algorithm from \citet{Biswas2021} for sampling from the global optimum of a discrete Boltzmann distribution $p = 1/Z \exp{(-y/T)}$ with temperature $T$ and $y = -\log p(x)$.
    We apply this to sample from our product of experts ( \Eqref{eq:PoE}).
    At each sampling step, a new sequence is proposed by randomly sampling $m \sim \text{Poisson}(\mu - 1)+1$ uniformly sampled amino acid substitutions with rate $\mu \sim \text{Uniform}(1,2.5)$.
    Proposed variants are accepted with probability $\min(1, \exp( (\hat{y} - y) / T))$ with $T$ annealed over time to encourage the sampler to become more exploitative.
    Variants with higher $\log p(x)$ are always accepted whereas sequences that decrease $\log p(x)$ are accepted with a probability proportional to the difference scaled by the temperature.
    \item \textbf{Random sampling:} We provide a simple baseline that does \emph{not} use search but rather uniformly samples variants around the WT protein, with aims of quantifying the advantage of using search to accumulate multiple promising mutations.
    Given a budget of $N$ variants, we use the same mutation proposal algorithm as \texttt{simulated annealing} to sample and apply a single mutation (which could consist of multiple amino acid substitutions) to the WT.
    After sampling $N$ such variants, we sort them by $\log p(x)$ and return the top $K$.
    \item \textbf{MALA-\emph{approx}:} This baseline employs a continuous relaxation of discrete protein space to use stochastic gradient-based continuous optimization inspired by~\citet{Nguyen2017}.
    During each step, \texttt{MALA-\emph{approx}} samples 
    \begin{equation}
    \Tilde{x}' \sim \mathcal{N}(\Tilde{x} + \frac{\epsilon}{2}\nabla_{\Tilde{x}} \log p(\texttt{round}(\Tilde{x})), \epsilon^2),        
    \end{equation} where $\epsilon$ is a step size and $\nabla_{\Tilde{x}} \log p(\texttt{round}(\Tilde{x}))$ is a straight-through estimate of the gradient~\citep{bengio2013estimating} of the product of experts with respect to the current relaxed variant.
    The continuous relaxation is based on a temperature-controlled relaxation of categorical variables to continuous vectors on the simplex~\citep{jang2016categorical,maddison2016concrete} with temperature $\tau$. 
    That is, we interpret each sequence position of protein $x$ to be a rounded sample from a relaxed categorical distribution.
    We run this sampler in ``logit space'' of these distributions since the support of logit space extends across the entire real line, whereas enforcing valid probability distributions at each sequence position would require solving a difficult constrained optimization problem.
    Given the one-hot WT protein $x^{WT}$, the sampler is initialized at logit values given by a convex combination of a uniform distribution over all amino acids and the wild type: $\log ( (1 - \tau) \cdot 1/|V| + \tau \cdot x^{WT} )$.
    To \texttt{round} to discrete values for evaluating $\log p(\texttt{round}(\Tilde{x}))$), we sample from the relaxed categorical distributions with logits $\Tilde{x}$ and then take the argmax at each sequence position.
    \item \textbf{CMA-ES:} The \texttt{CMA-ES} algorithm~\citep{hansen1996adapting} is a powerful evolutionary strategies optimizer that accelerates search by maintaining an estimate of higher order relational statistics between each dimension of the search space.
    This algorithm is designed for continuous spaces with tens or hundreds of dimensions to eventually concentrate around the global optimum.
    It iteratively updates a Gaussian distribution (both mean and full covariance matrix) by sampling and evaluating a population of candidates.
    Like \texttt{MALA-\emph{approx}}, it requires using a continuous relaxation of protein space.
    Simply, we ``flatten'' a one-hot protein $x$ into a vector of dimension $LV$ which we then re-interpret as a continuous vector in $\mathbb{R}^{LV}$.
    To undo this relaxation, we reshape the vector into a length $L$ sequence and interpret the $V$ entries at each sequence position as scores, i.e., we take the argmax.
\end{itemize}
We use multiple metrics to compare samplers that measure not only activity but also whether designed proteins are evolutionarily plausible.
Our results provide strong evidence that combining unsupervised evolutionary models with supervised models produces variants more likely to appear in nature and with higher fitness, especially when using unsupervised models fit on aligned sequences.
Moreover, we find that composing multiple unsupervised models trained on aligned \emph{and} unaligned sequences has the best performance.
We also verify that, by and large, our gradient-based discrete MCMC sampler for product of experts achieves better sampling performance than the random-walk-based and evolutionary-strategies-based baselines.

\subsection{MNIST-Sum}
\begin{wrapfigure}[15]{r}{0.3\textwidth}
    \vspace{-2mm}
    \centering
    \includegraphics[width=0.26\textwidth]{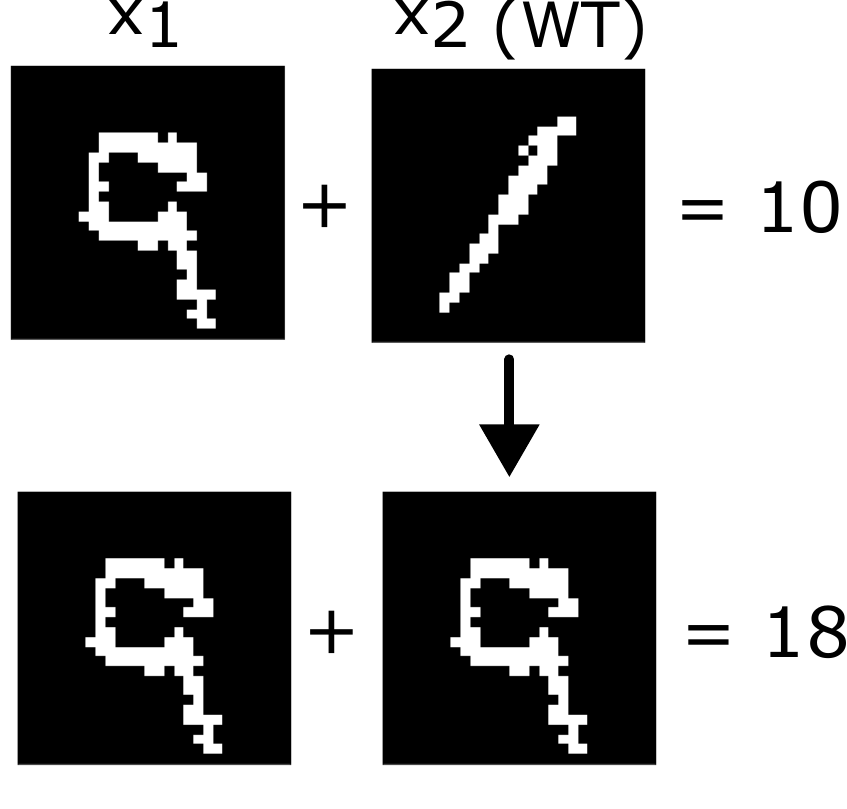}
    \caption{\textbf{MNIST-Sum}: The objective is to evolve the  wild type image (right) to maximize the sum of the two digits. \label{fig:mnist-sum}}
\end{wrapfigure}

Before we present results on ML-based directed evolution, we evaluate the samplers on a toy search problem where optimized designs are easy to interpret.

\textbf{Setup: }We construct a binary MNIST task that emulates a typical ML-based directed evolution problem. We are given two binary MNIST images $x_1$ and $x_2$ where $x_2$ is the wild type image.
Each image is treated as a length $L=784$ one-hot sequence where $V = \{0,1\}$.
The goal is to evolve $x_2$, keeping $x_1$ fixed, to maximize the sum of the two digits (\Figref{fig:mnist-sum}). 
Here, the equivalent of an amino acid substitution is flipping a binary pixel.
This task is difficult for methods that optimize in discrete space, since flipping most pixels causes the image to ``fall off'' the underlying low dimensional manifold of MNIST digits.
To obtain supervised experts $p_g(x)$ for a product of experts $\log p(x)$ (\Eqref{eq:PoE}), we use a labeled dataset of 50K pairs of binary MNIST digits with sum $\leq 10$ (the ``assay-labeled proteins'') to train an ensemble of three shallow siamese ConvNets that regresses the sum $x_1 + x_2$. 
We then use an unlabeled datasest of 50K binary MNIST images (the ``unlabeled protein database'', such as UniRef) to train a deep EBM and to train a DAE for unsupervised experts $p_f(x)$ (see appendix for model details).

PPDE is compared against \texttt{simulated annealing}, \texttt{MALA-\emph{approx}}, and \texttt{CMA-ES}.
All methods use a single EBM unsupervised expert and ensemble of three CNN-based supervised experts.
To examine the importance of the unsupervised expert, we run PPDE with only supervised experts (\texttt{PPDE (None)} or \texttt{PPDE (supervised only)}).
We then highlight PPDE's plug and play ability by running it with an EBM expert (\texttt{PPDE (EBM)}) and then swapping it out for a DAE expert (\texttt{PPDE (DAE)})\emph{ without performing any further fine-tuning or re-training}.
Key hyperparameters for each sampler are tuned on the digit pair $x_1=1, x_2=5$ (which has maximum sum $10$ since $x_2$ is evolved while $x_1$ is held fixed) for $10K$ steps.
We test each sampler on the out-of-distribution image pair $x_1=9, x_2=1$ with maximum possible sum 18.
In this case, ``solving'' the task is equivalent to evolving the wild type ``1'' digit into a ``9''.
Each method evolves a population of size 128 and each sample in the population is optimized for a budget of $20K$ steps.
To estimate ``ground truth'' sums for optimized image pairs, we train an ensemble of three siamese ConvNets  to near perfect accuracy  on pairs of digits whose sums are $\leq 18$. 

\begin{figure}
    \centering
    \begin{subfigure}{0.32\textwidth}
        \includegraphics[width=\textwidth]{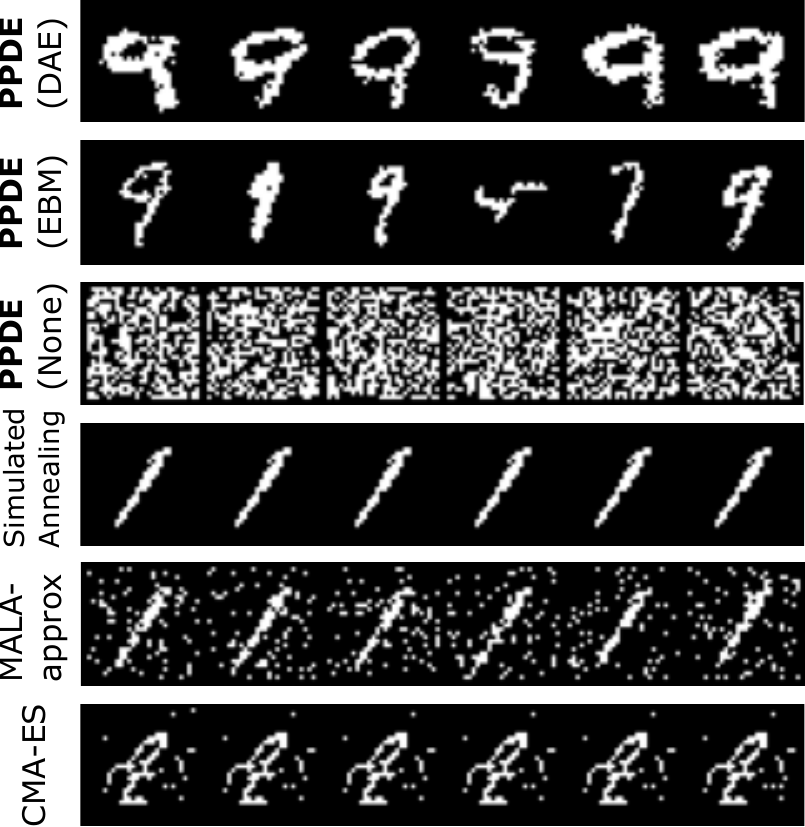}
        \caption{}
        \label{fig:mnist-vis}
    \end{subfigure}
    \begin{subfigure}{.67\textwidth}
        \includegraphics[width=\textwidth]{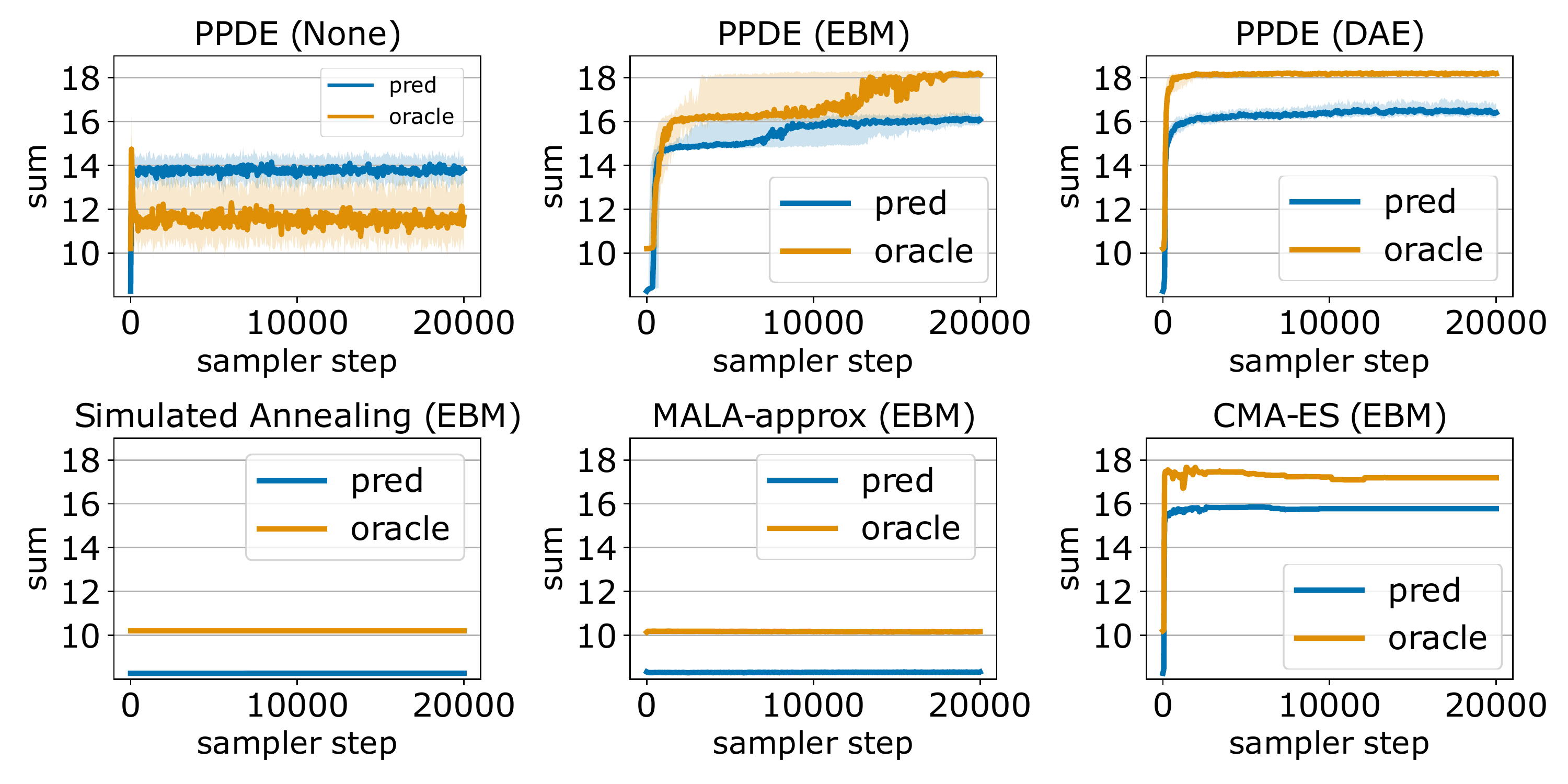}
        \caption{}
        \label{fig:mnist-scores}
    \end{subfigure}
    \caption{\textbf{MNIST-Sum results}: a) Random samples from the optimized digit populations. The samples from \texttt{PPDE (DAE)} have the highest diversity and also closely resemble MNIST images of a ``9''. \texttt{PPDE (None)} does not use an unsupervised expert which causes sampler trajectories to become trapped in adversarial regions of the space where samples resemble white noise. The random mutation proposals employed by \texttt{simulated annealing} and \texttt{MALA-\emph{approx}} cause these samplers to fail due to the high-dimensionality of the space. \texttt{CMA-ES}'s samples have low diversity yet converge onto a population of images that somewhat resemble a ``9''.  b) The supervised experts' predicted sum (orange) vs. the oracle's predicted sum (blue) for the sampler trajectories. Each line plot is the 70\textsuperscript{th} percentile sum with the shaded region showing the 50\textsuperscript{th} and 90\textsuperscript{th} percentiles. \texttt{PPDE (DAE)} reliably maximizes the sum of the digits (``18'') as measured by the oracle model.}
    \label{fig:mnist}
\end{figure}
\textbf{Results:}  
The qualitative and quantitative results in \Figref{fig:mnist} show that composing the supervised experts with an unsupervised expert (the EBM or the DAE) is \emph{necessary} to solve the task; otherwise, the PPDE sampler is unable to avoid adversarial inputs and each run of the sampler converges to white noise images.
We also observe that PPDE samples a highly diverse population of digits with noticeable semantic differences.
We see that  \texttt{PPDE (DAE)} outperforms \texttt{PPDE (EBM)}, which we suspect could be caused by the DAE's gradients being more informative due to its denoising training objective.
\texttt{Simulated annealing} and \texttt{MALA-\emph{approx}} fail in this high-dimensional space due to the difficulty of using random walks to find search directions that increase the product of experts probability.
\texttt{CMA-ES} is outperformed by PPDE despite its use of covariance information to accelerate search in high-dimensional spaces. 
PPDE's evolved images are considerably more realistic than those produced by CMA-ES.
We also observe that on this task, \texttt{CMA-ES} produces images that show low \emph{semantic} diversity (the images differ at the pixel-level but the by and large look highly similar).
We speculate that PPDE gains an advantage over CMA-ES by using gradients to select which pixels to mutate instead of approximate high-order statistics.

\subsection{\emph{In silico} Directed Evolution}
\textbf{Datasets: }We use three benchmark proteins with partially characterized wide fitness landscapes and MSAs from~\citet{hsu2022learning} for \emph{in silico} directed evolution experiments: the Poly(A)-binding protein (PABP) dataset of variants measuring binding activity (95 residues, each variant has $\leq$ 2 mutations), the ubiquitination factor E4B (UBE4B) protein dataset measuring ligase activity (103 residues, each variant has $\leq$ 6 mutations), and GFP protein dataset measuring fluorescence (238 residues, each variant has $\leq$ 15 mutations).
To emulate a realistic protein engineering setup with reasonable amounts of data for training our supervised experts $g(x)$, we use the ``2-vs-rest'' mutation train/test split as suggested in~\citet{dallago2021flip}---that is, after splitting each dataset 80/20, we  keep only sequences with two or fewer mutations for training and subsample 10\% of these. %
For PABP, both the train and test sets only have variants with at most 2 mutations.
This amounts to 3K sequences for PABP,  2K for UBE4B, and 1K for GFP.

\textbf{Metrics: }Quantitative evaluation  of optimized variants involves measuring the activity (log fitness),  the likelihood to appear in nature (evolutionary density), population diversity, and ability to explore the fitness landscape around the WT (average number of mutations per variant). 
To compute log fitness, we take inspiration from model-based optimization benchmarks~\citep{Trabucco2022} and simulate an expensive wet lab verification with a powerful ``oracle'' model that we train with 80\% of all variants (with all numbers of mutations) for each protein.
Log fitness is scored relative to WT with the Augmented EVmutation Potts model from~\citet{hsu2022learning} as the oracle.
To compute an evolutionary density score relative to WT, we use a different transformer, the MSA Transformer~\citep{rao2021msa}, conditioned on 500 randomly subsampled sequences from each MSA.
For both log fitness and evolutionary density, positive values mean the variant outscored the WT while negative values are worse than WT.
We run each sampler 128 times for 10K steps each, initialized at the WT, and select the sample with the maximum $\log p(x)$ per run for the final population.
As plug and play baselines we again use \texttt{simulated annealing}, \texttt{CMA-ES}, and \texttt{MALA-\emph{approx}}, as well as \texttt{random sampling}.

\textbf{Experts: }We use the EVmutation Potts and ESM2 family of protein language models as pre-trained unsupervised experts.
All baselines use a single Potts expert and are directly compared with \texttt{PPDE (Potts)}.
Unless stated otherwise, we use the 150M parameter version of ESM2.
We also run our sampler with \emph{both} Potts and ESM2 (i.e., we take the sum of the Potts and ESM2 evolutionary density scores) (\texttt{PPDE (Potts+ESM2)}).
For the supervised expert we use an ensemble of three shallow CNNs introduced as a baseline in~\citet{dallago2021flip} which regress activity directly from one-hot encoded proteins.
See the appendix for architecture details and details on how we implement the evolutionary density scoring for the Potts and protein language model scoring functions.

\textbf{Choosing $\lambda$: } Multiplying the supervised experts with the hyperparameter $\lambda \geq 0$ trades off how much the unsupervised and supervised experts influence which variants are sampled from the product of experts distribution.
A larger value for $\lambda$ inflates the supervised expert scores which causes the sampler to have more confidence in variants with high predicted activity.
We use a simple heuristic to select $\lambda$ for the three benchmark proteins and for each of the Potts and ESM2 unsupervised experts.
The heuristic roughly assigns equal emphasis to the unsupervised and supervised expert scores.
In detail, we estimate the \{\texttt{min}, \texttt{max}, \texttt{mean}\} statistics of expert scores computed from a small representative sample of labeled protein variants and then picks $\lambda$ such that the statistics for the supervised expert scores are close to those of the unsupervised expert scores.
We sample 100 protein variants from the training set that have lower activity than WT and 100 proteins with higher activity higher than WT and then evaluate the \{\texttt{min}, \texttt{max}, \texttt{mean}\} of the unsupervised expert scores and $\lambda \times $\{\texttt{min}, \texttt{max}, \texttt{mean}\} of the supervised expert scores for $\lambda \in \{0.5, 1, 3,5, 15\}$.
When using the Potts unsupervised expert we use $\lambda = 5, 0.5, 15$ and when using the ESM2 expert we use $\lambda = 5,3, 1$ for the PABP, UBE4B, and GFP proteins, respectively.
Less manual approaches that tune $\lambda$ over a grid of values or \emph{learn} $\lambda$ over a held-out validation set of labeled variants could be used instead if desired~\citep{holtzman2018learning}.

\begin{figure}[t]
    \centering
    \includegraphics[width=\textwidth]{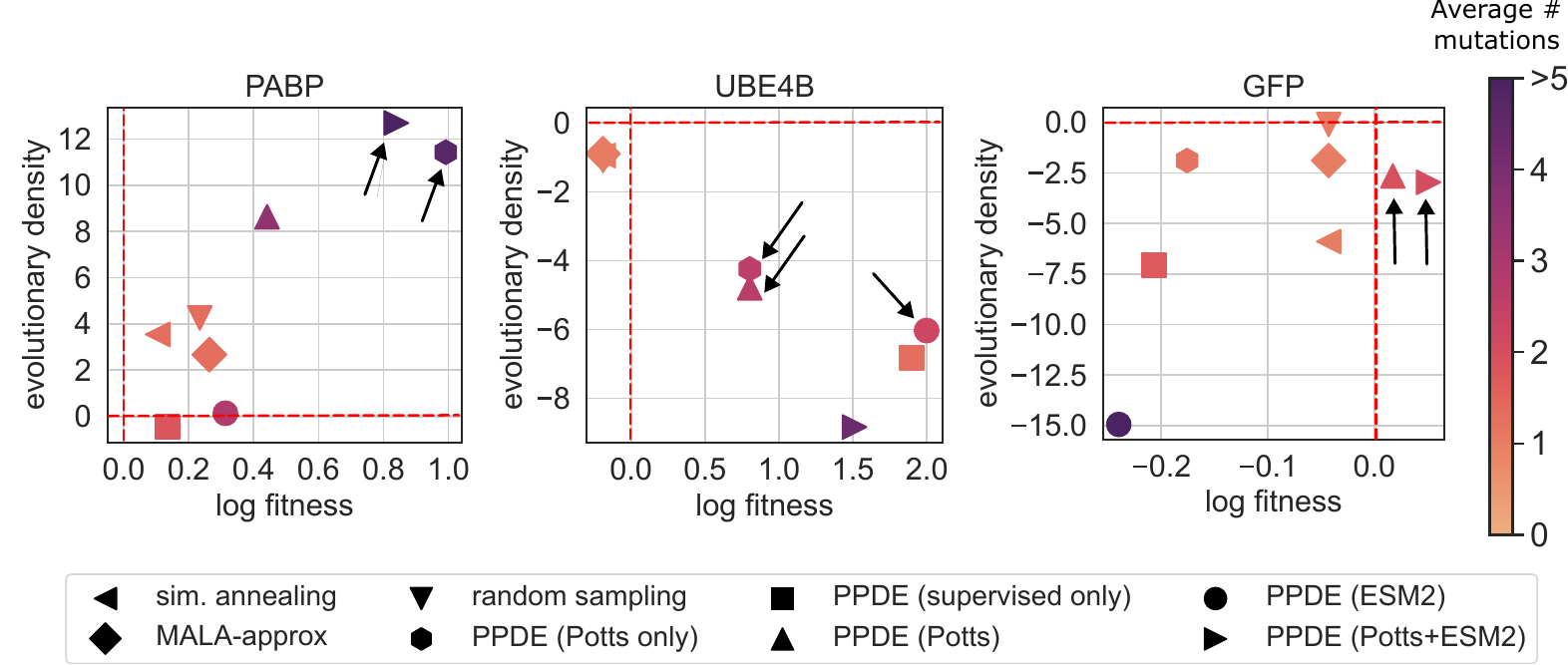}
    \caption{\textbf{\emph{In silico} directed evolution 80\textsuperscript{th} percentile results}. Positive log fitness and evolutionary (evo.) density scores are better than the WT while negative values are worse than WT. The red dashed lines intersect at WT performance $(0,0)$.
    Each sampler is colored by the average number of mutations per variant (higher is better).
    On PABP (left), \texttt{PPDE (Potts+ESM2)} and \texttt{PPDE (Potts only)} (arrows) achieve the highest evo. density and log fitness---both better than WT---and with relatively high numbers of mutations.
    On UBE4B (center),  \texttt{PPDE (Potts)} and \texttt{PPDE (Potts only)} (arrows) appear to best trade off evo. density, log fitness, and mutation count. \texttt{PPDE (ESM2)} achieves the best log fitness but with slightly worse evo. density scores.
    On GFP (right), only \texttt{PPDE (Potts)} and \texttt{PPDE (Potts+ESM2)} (arrows) discover samples with log fitness better than WT.
    \texttt{CMA-ES} is omitted for viewing clarity due to outlier results on UBE4B (log fitness: $2.54$, evo. density: $\mathbf{-94.76}$) and GFP (log fitness: $-1.13$, evo. density: $\mathbf{-22.85}$). Best viewed in color.}
    \label{fig:protein_quant}
\end{figure}

\begin{table}[t]
\caption{\textbf{Population diversity} (\% of variants that are unique).\label{tab:diversity}}
\centering
\begin{adjustbox}{max width=\textwidth}
\begin{tabular}{@{}lccccccccc@{}}
\toprule
 & \begin{tabular}[c]{@{}c@{}}Random\\search\end{tabular} & \begin{tabular}[c]{@{}c@{}}Simulated\\annealing\end{tabular} & MALA\emph{-approx} & CMA-ES  &
 \begin{tabular}[c]{@{}c@{}}PPDE\\(Potts only)\end{tabular} &
 \begin{tabular}[c]{@{}c@{}}PPDE\\(Super. only)\end{tabular} & \begin{tabular}[c]{@{}c@{}}PPDE\\(Potts)\end{tabular} & \begin{tabular}[c]{@{}c@{}}PPDE\\(ESM2)\end{tabular} & \begin{tabular}[c]{@{}c@{}}PPDE\\(Potts+ESM2)\end{tabular} \\ \midrule 
PABP & 32.8 & 28.9 & 28.9&  0.8 & \textbf{85.2} & 60.2 & 65.6 & 63.1 & \textbf{85.2} \\
UBE4B & 7.0 & 4.7 & 6.2 & 3.1 & 12.5 & 18.8 & 18.8 &  \textbf{36.5} & 31.3 \\
GFP & 9.4 & 3.9 & 9.4  & 92.2 & 8.6 & 59.4 & 22.7 & \textbf{92.9} & 21.9 \\
\bottomrule
\end{tabular}
\end{adjustbox}
\end{table}

\begin{figure}[t]
     \begin{subfigure}{0.32\textwidth}
           \centering
        \includegraphics[width=\textwidth]{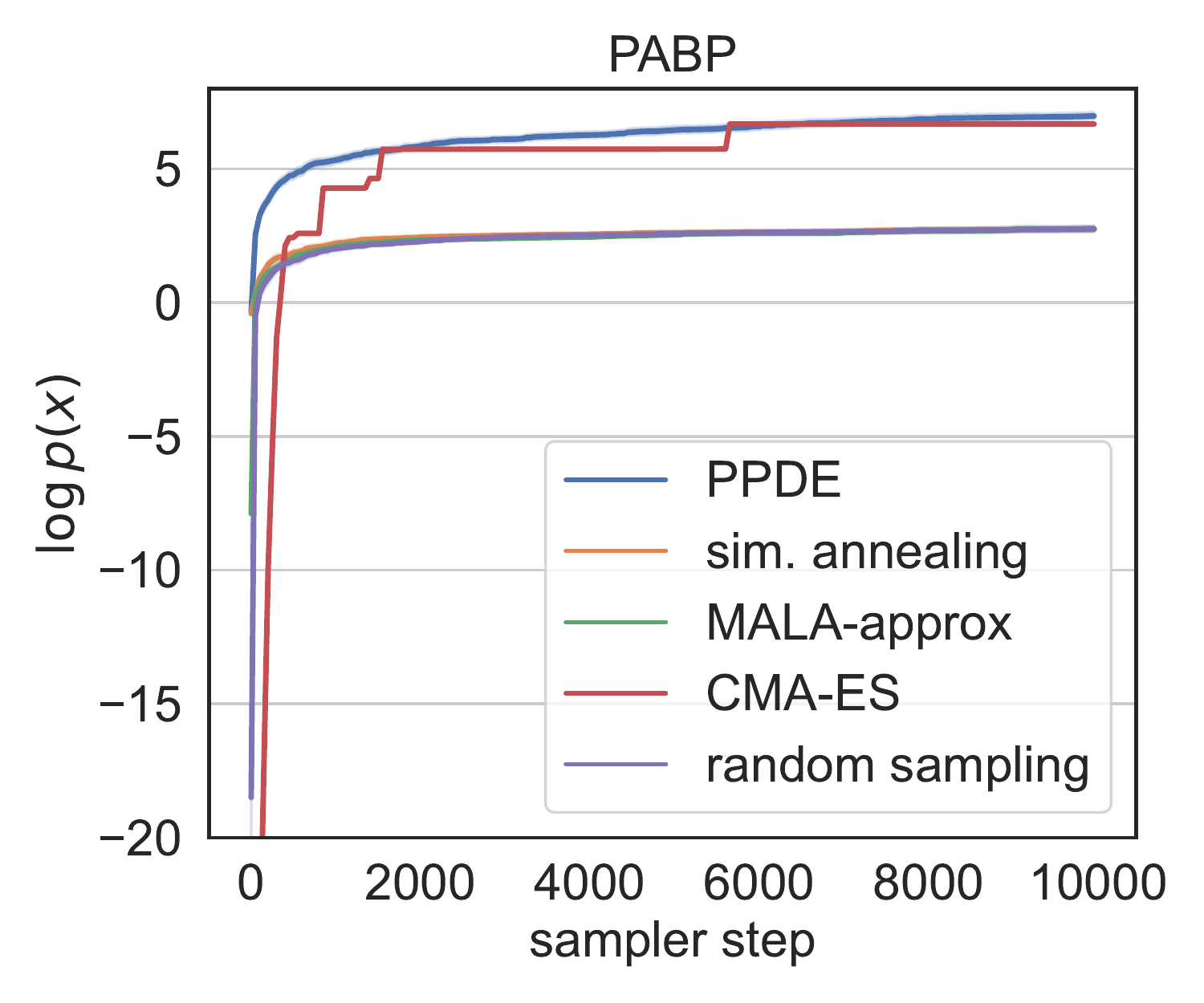}
        \caption{}
        \label{fig:efficiency_pabp}
    \end{subfigure}
    \begin{subfigure}{0.32\textwidth}
           \centering
        \includegraphics[width=\textwidth]{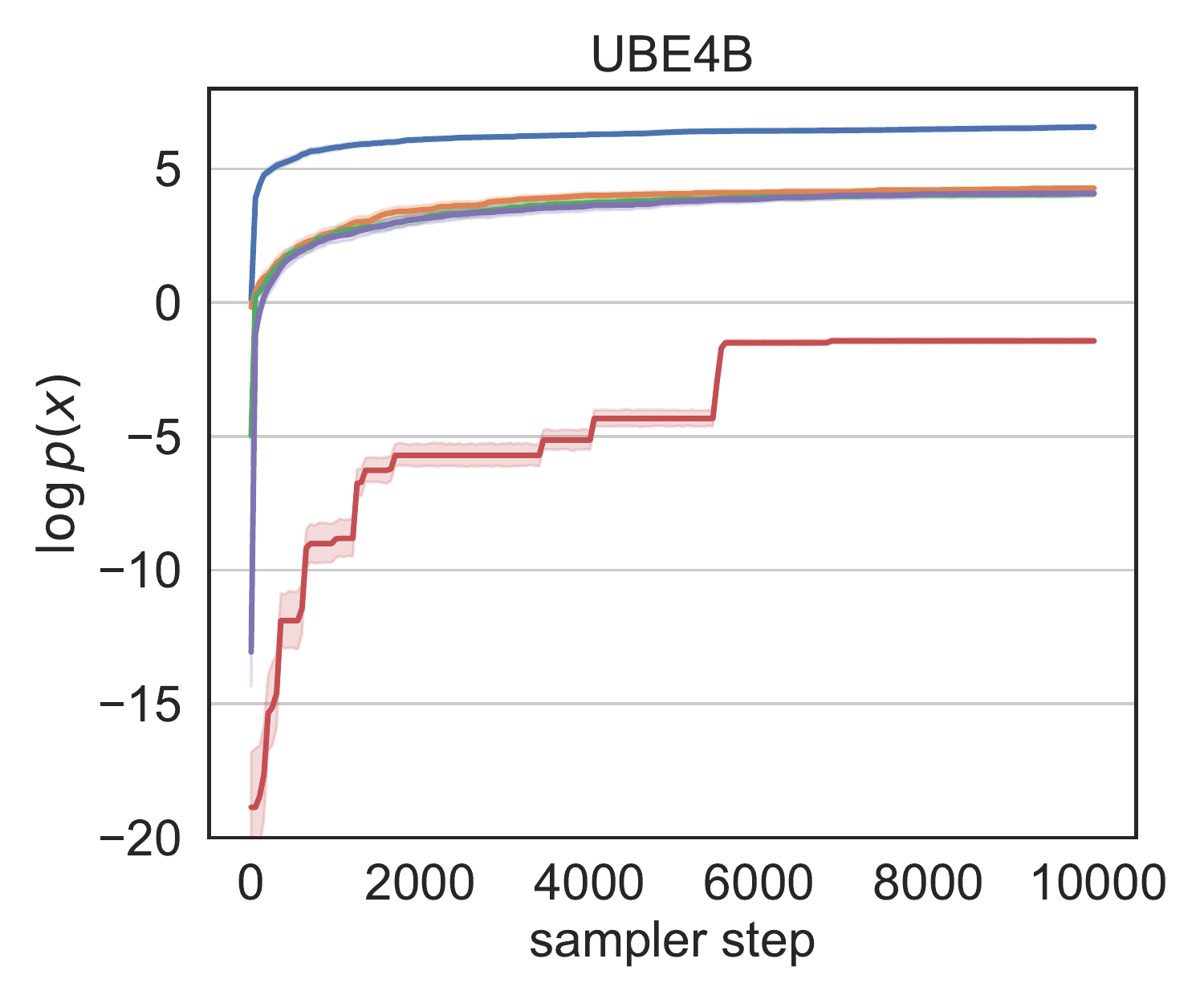}
        \caption{}
        \label{fig:efficiency_ube4b}
    \end{subfigure}
    \begin{subfigure}{0.32\textwidth}
           \centering
        \includegraphics[width=\textwidth]{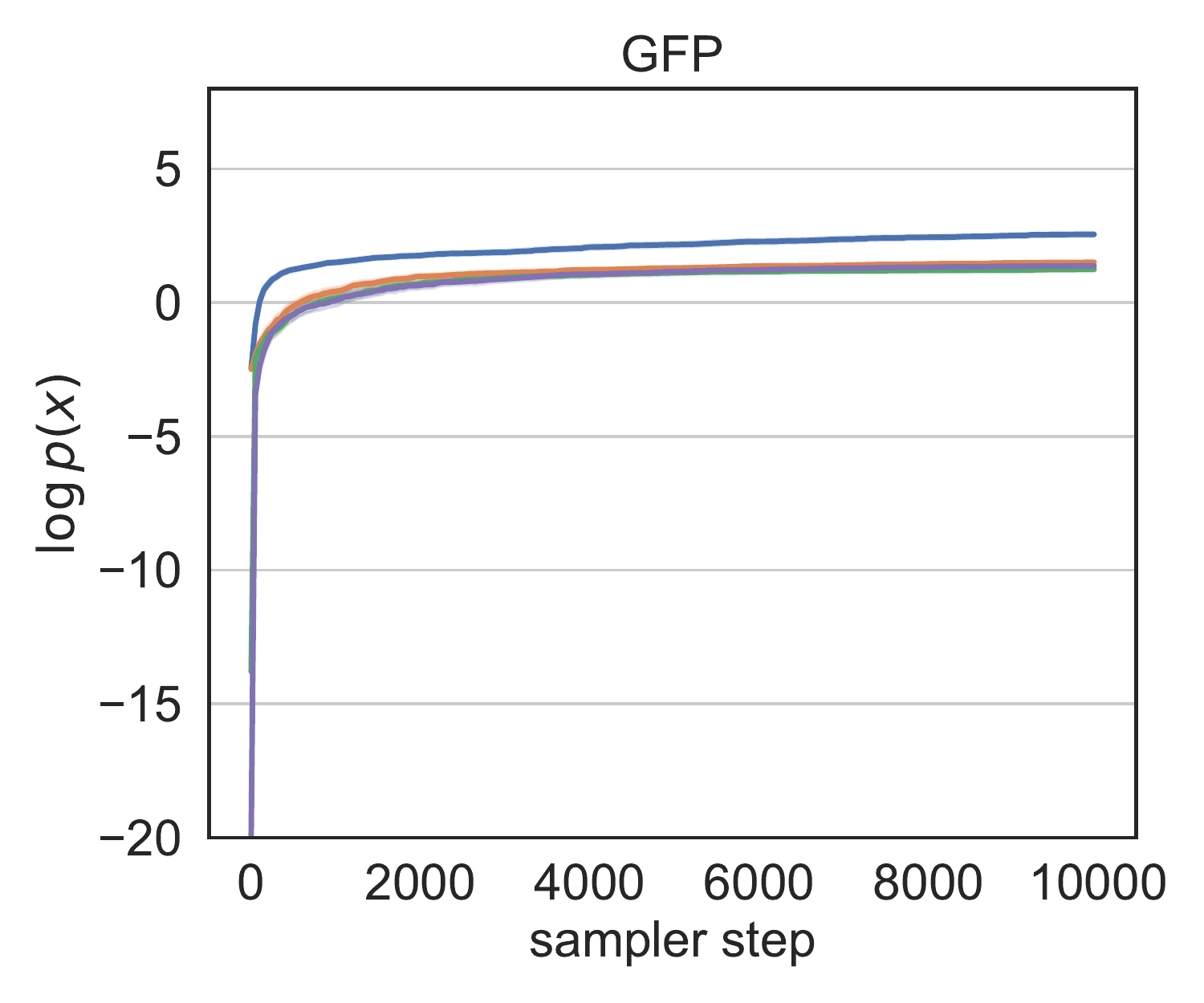}
        \caption{}
        \label{fig:efficiency_gfp}
    \end{subfigure}
    \caption{\textbf{PPDE is most efficient at finding good optima of the high-dimensional product of experts.} Cumulative maximum product of experts log probability averaged across the population (higher is better). All samplers use the same product of experts target distribution with Potts unsupervised expert (other unsupervised experts not shown for clarity of presentation). \texttt{CMA-ES} is not visible in c) because the optimization starts and remains around $\log p(x) \approx$ -50.\label{fig:efficiency}} 
\end{figure}
\subsubsection{Main results}
\Figref{fig:protein_quant} shows the 80\textsuperscript{th} percentile metrics for each evolved population colored by the population's average number of mutations.
 The 50\textsuperscript{th} and 100\textsuperscript{th} percentile results for all samplers are provided in a table in the appendix (Table~\ref{tab:main}).
 Diversity scores are in Table~\ref{tab:diversity}.
 We make the following observations.

 \textbf{Diversity: }Across all three proteins, PPDE-based samplers achieve the best diversity (highest percentage of unique variants), implying superior exploration of good optima around the WT protein. 
 
 \textbf{Evolutionary density, log fitness, and mutation counts: }Out of the samplers using the Potts unsupervised expert, \texttt{PPDE (Potts)} discovers variants with higher log fitness and average number of mutations than \texttt{random search}, \texttt{simulated annealing}, and \texttt{MALA-\emph{approx}}.
In terms of evolutionary density, \texttt{PPDE (Potts)} achieves the highest scores on PABP, and on the more challenging UBE4B and GFP proteins, we suspect that the slightly lower 80\textsuperscript{th} percentile evolutionary density scores compared to the baselines (except for \texttt{CMA-ES}, which scored poorly) are in part due to PPDE discovering variants with higher average mutations ($2-4+$ mutations vs. $\sim 1$ mutation for the baselines) and higher log fitness.
We found that the \texttt{CMA-ES} sampler had inconsistent performance across the three proteins.
Among samplers using the Potts unsupervised expert, it achieves the highest fitness scores on two proteins (PABP and UBE4B), but does so with diversity scores of 0.8\% and 3.1\% compared to 85.2\% and 12.5\% for \texttt{PPDE (Potts)} (Table~\ref{tab:diversity}).
CMA-ES finds variants with high numbers of mutations ($\sim 10-17$) but low evolutionary density scores (50\textsuperscript{th} percentile scores of 3.47 on PABP, -94.76 on UBE4B and -62.43 on GFP compared to 6.92, -4.79, and -5.98 for PPDE; see Table~\ref{tab:main}).
It finds just one protein variant for PABP that is 17 mutations from WT with an evolutionary density score of 3.47, and it seems to have significant difficulty with the larger proteins UBE4B and GFP; e.g., on GFP the average 50\textsuperscript{th} percentile log fitness is -2.50 compared to -0.04 for PPDE.
\Figref{fig:efficiency} shows sampler trajectories, with which we see that across all three proteins, PPDE is the superior approach for efficiently sampling from the high-dimensional discrete product of experts distributions. 
We see here again that \texttt{CMA-ES} performs worse as the protein sequence length and fitness landscape complexity increases.
It appears that a major limitation of CMA-ES is its continuous relaxation, which projects each sequence in the population to a variant far from the WT.
For GFP, the population average product of experts probability starts and stays around $\log p(x) \approx -50$.
We conclude that \texttt{CMA-ES} can be recommended for use only when a single protein variant is desired and the length of the protein is relatively small (e.g., $\leq$ 95 residues).

\textbf{Swapping unsupervised experts:} \emph{Without using an unsupervised expert}, PPDE discovers variants with worse evolutionary density scores than even \texttt{random sampling}, confirming our observations from the previous MNIST-Sum experiment that supervised experts alone are susceptible to adversarial inputs.
We also examined PPDE without any supervised expert (\texttt{PPDE (Potts only)}) by setting $\lambda = 0$.
The observation that \texttt{PPDE (Potts only)} achieves a higher log fitness on PABP and only slightly worse log fitness on UBE4B than \texttt{PPDE (Potts)} is not surprising, as recent evidence suggests unsupervised evolutionary density scores are (at least moderately) predictive of mutation effects~\citep{Meier2021,hsu2022learning,weinstein2022nonidentifiability}.
However, the sampler performance degrades significantly by all metrics on GFP.
When using the ESM2 unsupervised expert, PPDE discovers variants with lower evolutionary density scores than when using the Potts.
This is not altogether unexpected since the MSA Transformer, which is used to score the variants, is conditioned on the same MSA as used to fit the Potts model.
Although, this result also suggests that using unsupervised experts fit to \emph{aligned sequences} is more promising for ML-based directed evolution.
We observe a promising compositional effect from combining unsupervised experts trained on unaligned sequences (ESM2) and aligned sequences (Potts); e.g., across all three proteins the \texttt{PPDE (Potts+ESM2)} sampler achieves higher 100\textsuperscript{th} percentile fitness and density scores than when using the Potts or ESM2 experts alone, and increases the diversity and average number of mutations as well. 

\subsubsection{Qualitative results}
\begin{figure}[t]
    \centering
    \includegraphics[width=\textwidth]{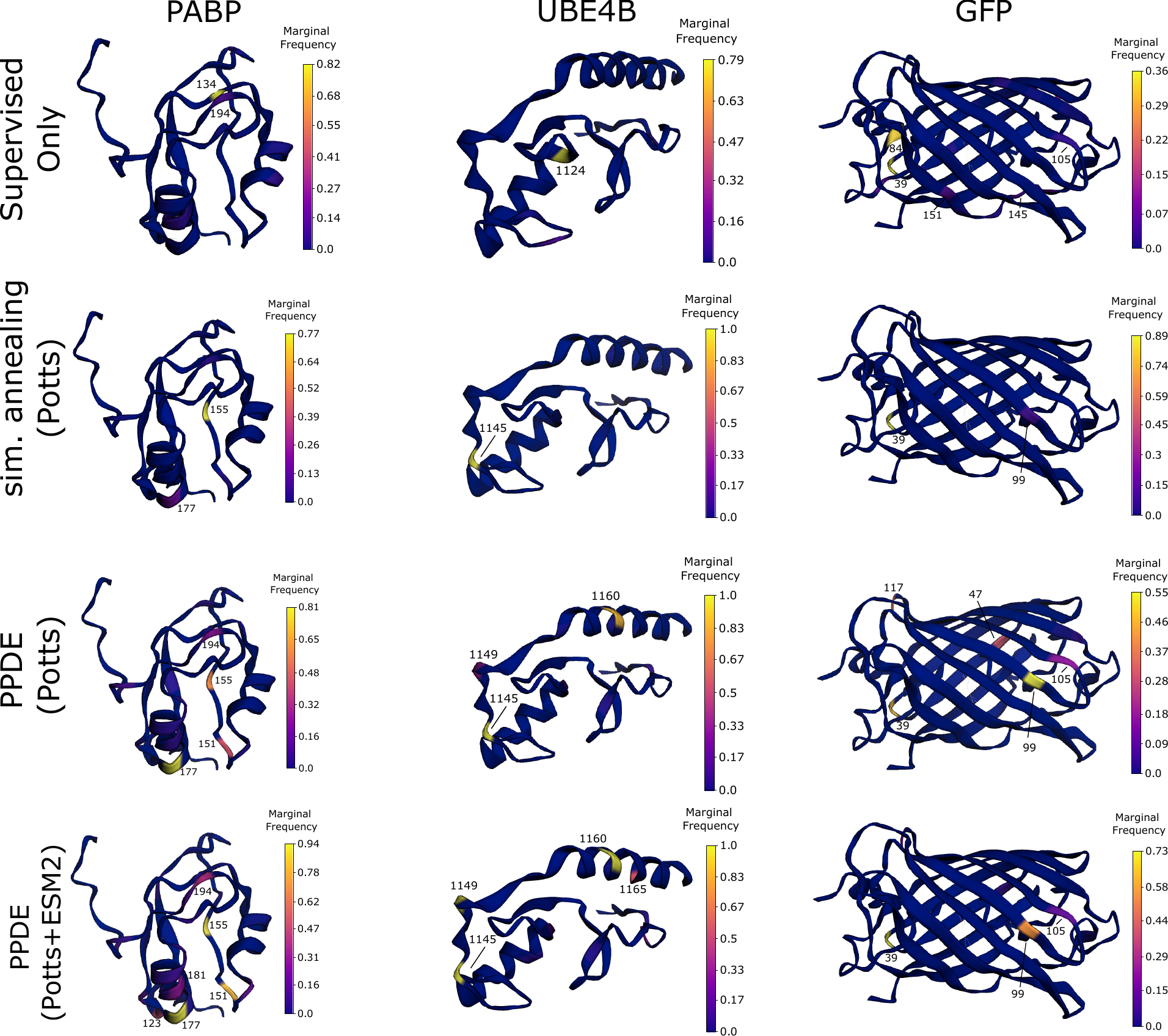}
    \caption{\textbf{Visualization of mutation site frequency}. Best viewed in color and zoomed in. The marginal frequency of a mutation site is the fraction of protein variants in the population with a mutation at a particular sequence position. We visualize site frequencies as a heatmap on the 3D structure of the WT protein. For example, nearly 100\% of UBE4B variants discovered by \texttt{PPDE (Potts+ESM2)} have a mutation at position 1145. We annotate only a few high frequency sequence positions per protein for clarity; many sequence positions with a low frequency are not annotated.}
    \label{fig:proteins_qual}
\end{figure}
We provide a qualitative analysis of the mutations discovered by the \texttt{supervised only}, \texttt{simulated annealing}, \texttt{PPDE (Potts)} and \texttt{PPDE (Potts+ESM2)} samplers in \Figref{fig:proteins_qual}.
Many of the variants found by PPDE tend to have the same 1-3 ``highly plausible'' mutations in addition to a few ``rare'' mutations that show up in variants at a much lower rate.

\subsubsection{Comparing protein language model experts}
\label{sec:swap_gen}
\begin{figure}
    \centering
    \includegraphics[width=\textwidth]{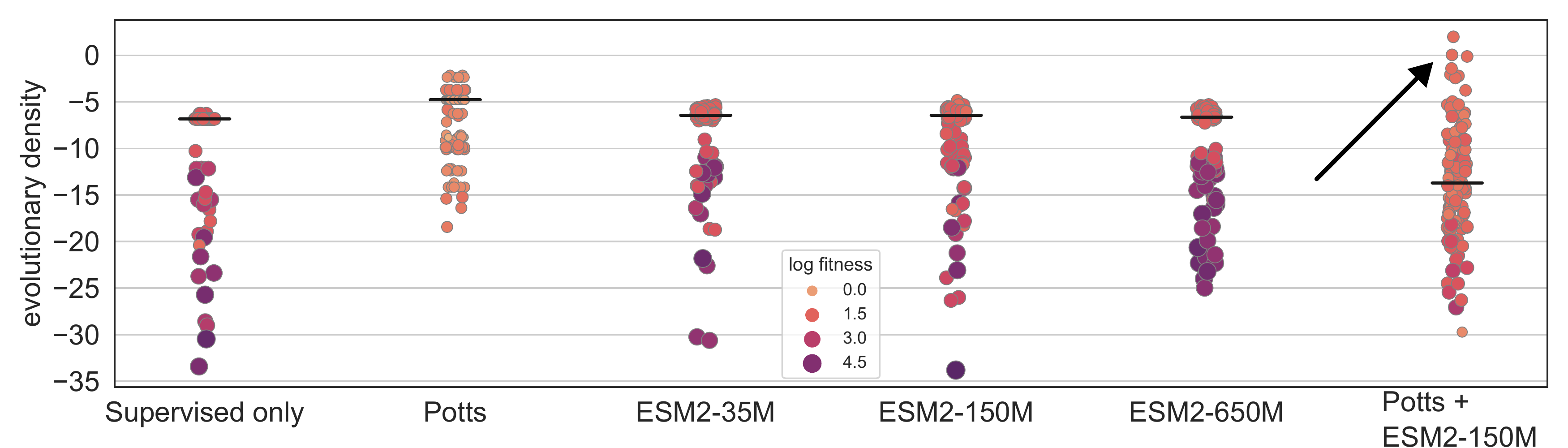}
    \caption{\textbf{Swapping in different protein language model experts}. Full populations (128 variants) colored by log fitness with the median evolutionary density annotated. We swap out different unsupervised experts for the \emph{in silico} directed evolution of UBE4B using the same supervised expert.
    This includes a 650M parameter protein language model (\texttt{ESM2-650M}).  Combining unsupervised experts trained on aligned and unaligned sequences (\texttt{Potts+ESM2-150M}) discovers variants with highest density and higher predicted fitness than WT (arrow).}
    \label{fig:priors}
\end{figure}
\Figref{fig:priors} highlights the ability of PPDE to successfully sample from the 35M, 150M, and 650M parameter ESM2 language models without any extra fine-tuning or re-training.
This flexibility makes it easy to leverage continued improvement in protein language models, or use language models fine-tuned for specific protein families.
As the number of ESM2 parameters increase, we observe a trend where variants with higher predicted activity and marginally higher evolutionary density are discovered.
A similar observation in~\citet{Nijkamp2022} also suggests that model size positively correlates with zero-shot fitness prediction on wide fitness landscapes.
When directly comparing \texttt{supervised only} and \texttt{PPDE (Potts)}, we can see a large improvement in evolutionary density when the Potts expert is used.
ESM2, which is trained on unaligned proteins, achieves lower evolutionary density scores than the Potts, suggesting it acts as a much softer search constraint.
Combining the Potts and ESM2 experts results in improving the top percentile of discovered variants compared to those found by using the Potts or ESM2 experts alone.

\subsubsection{Path length sensitivity analysis}
\begin{figure}[t]
    \centering
    \begin{subfigure}{.5\textwidth}
        \centering
        \includegraphics[scale=0.35]{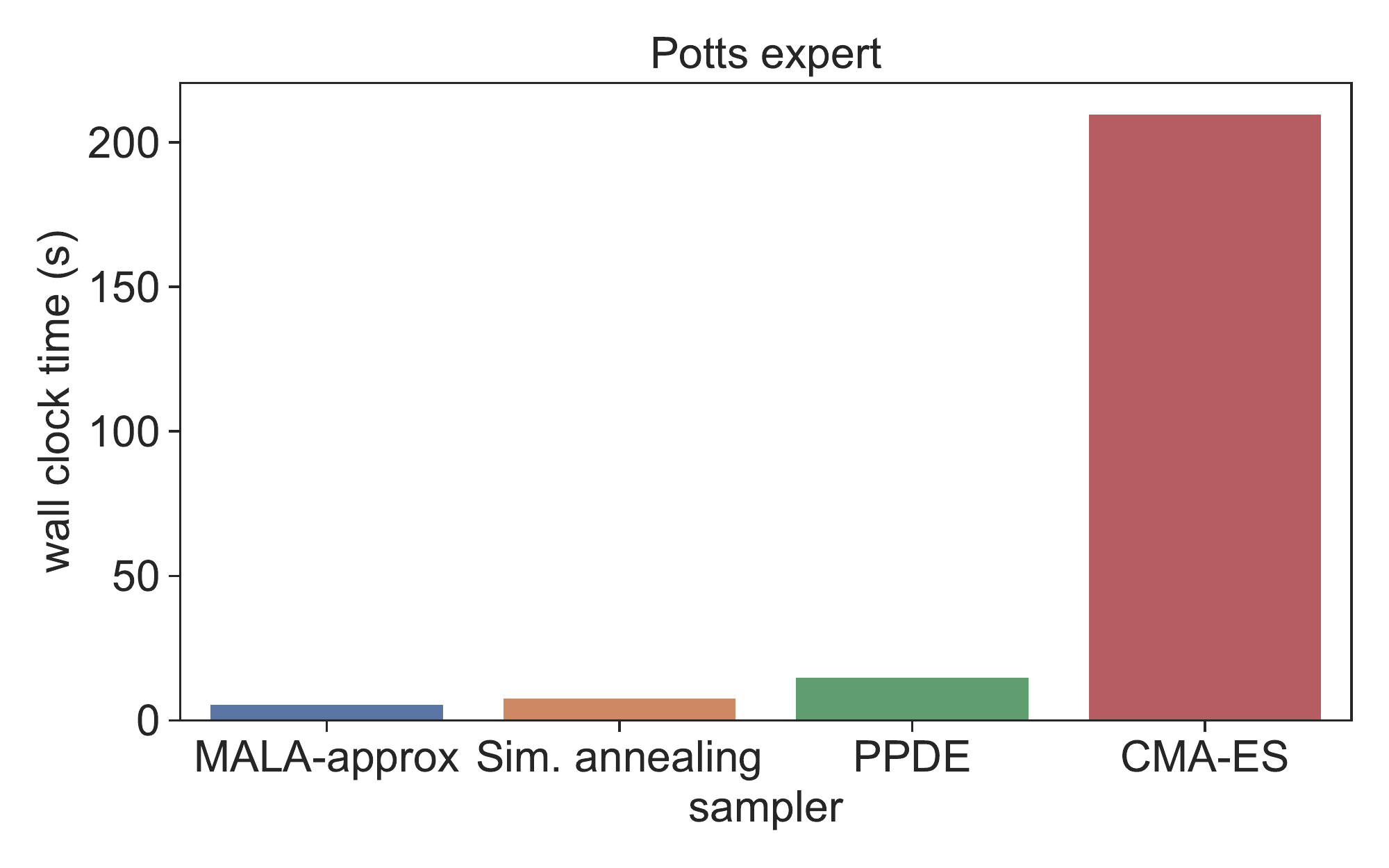}
        \caption{\label{fig:potts-compute}}
    \end{subfigure}%
    \begin{subfigure}{.5\textwidth}
        \centering
        \includegraphics[scale=0.35]{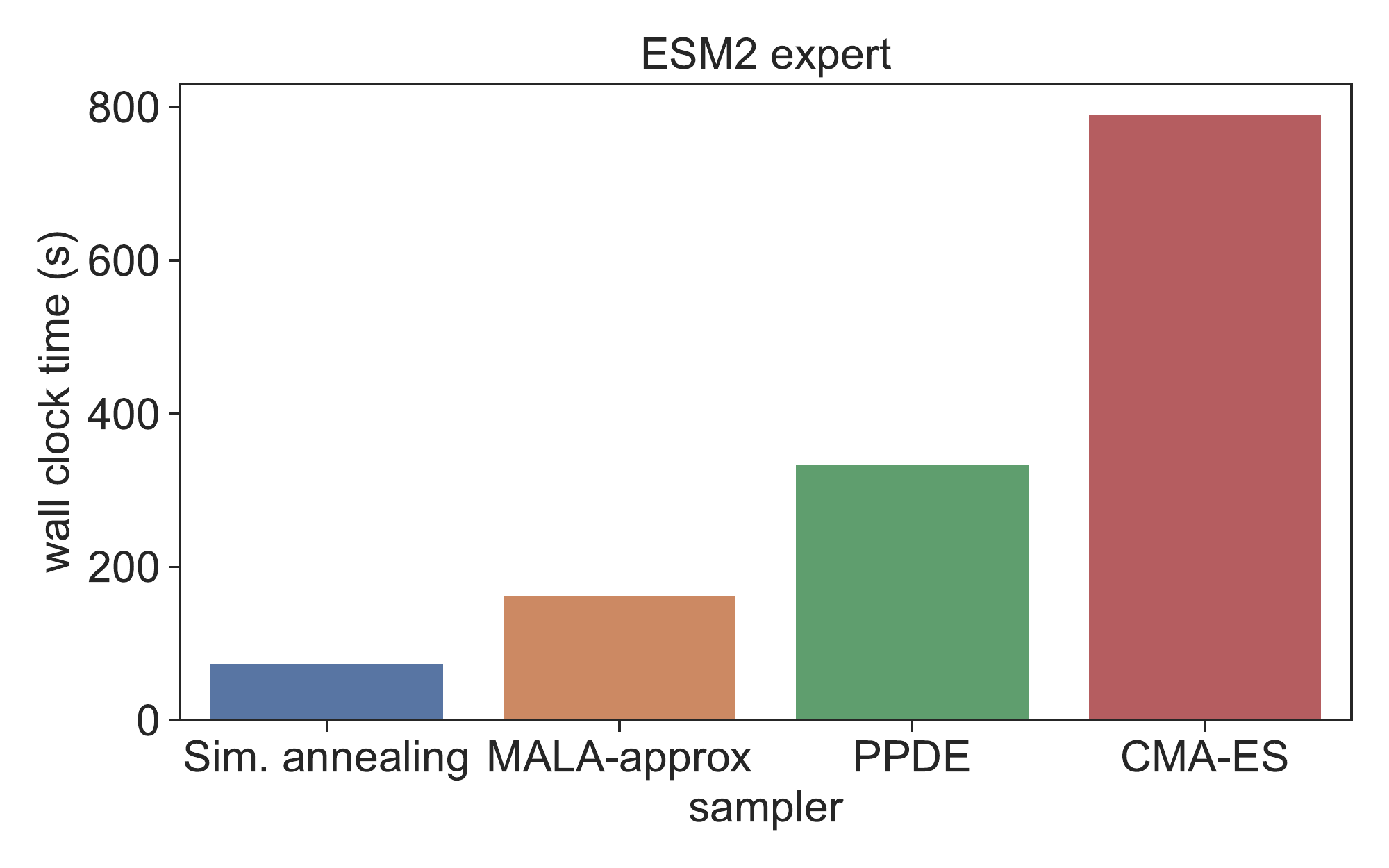}
        \caption{\label{fig:esm2-compute}}
    \end{subfigure}
    \begin{subfigure}{\textwidth}
        \centering
        \includegraphics[scale=0.4]{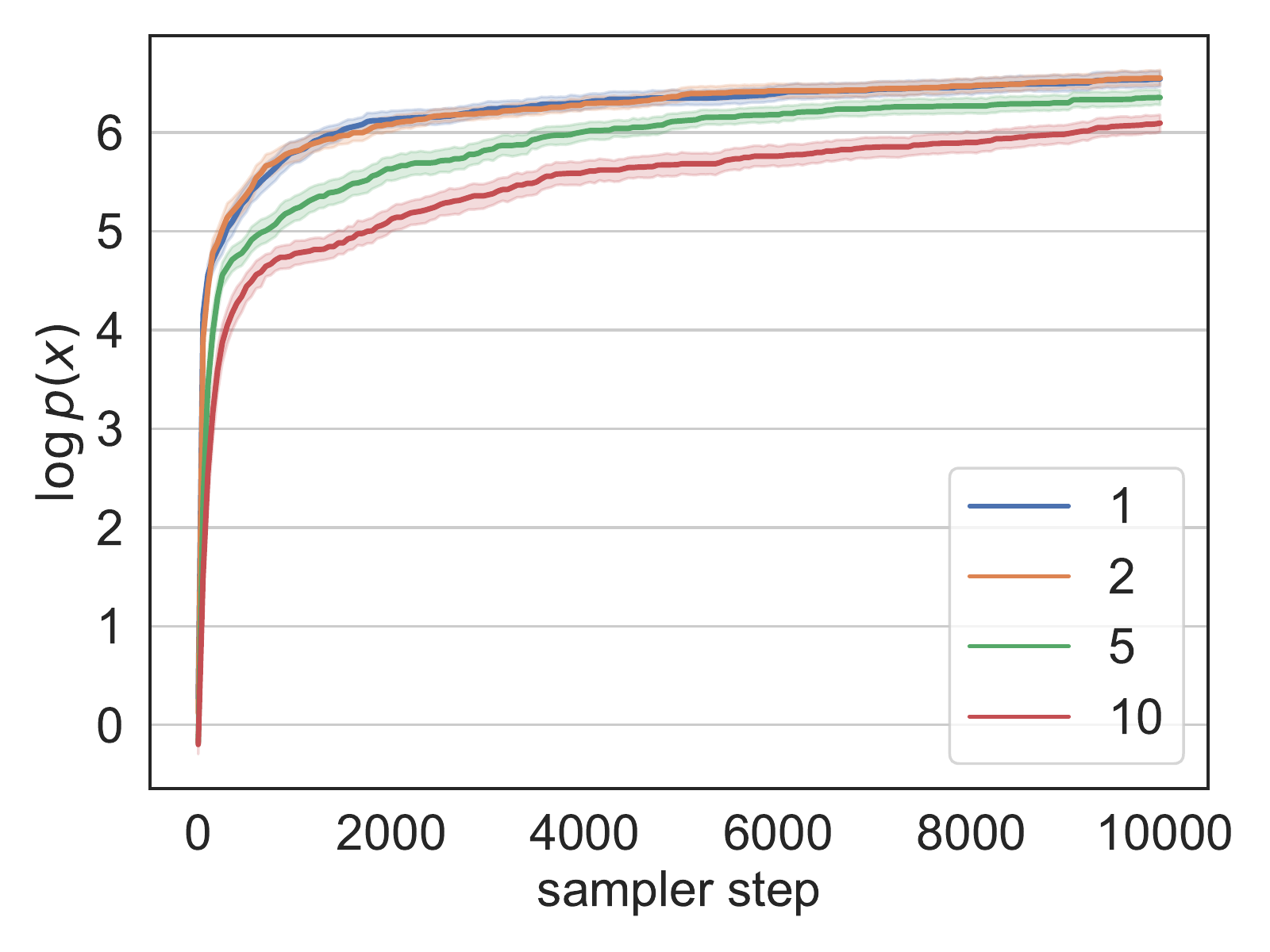}
         \caption{\label{fig:path-length}}
    \end{subfigure}   
    \caption{Wall clock run time comparisons using the Potts (a) and ESM2 (b) unsupervised experts. Run times are averaged over 7 trials using the PABP protein, 1K sampler steps, and a population of size 16. c) Comparing PPDE on UBE4B with various max path lengths $U = 2X-1$ where $X \in \{1,2,5,10\}$.\label{fig:sampler-analysis}}
\end{figure}
PPDE's proposal distribution samples a \emph{path} of $R \sim \text{Unif}(1,U)$ single substitutions at each sampler step to rapidly navigate through protein space.
We conducted a sensitivity analysis on the maximum path length $U$, where $U = 2X-1$ and $X \in \{1,2,5,10\}$ (\Figref{fig:path-length}) on the UBE4B protein. 
In this experiment, we found that path lengths longer than $3$ did not significantly improve sampler efficiency.
This is possibly because we already have a good initialization for the sampler (the WT) and short path lengths are sufficient for discovering high quality variants. 
It is also possible that, in our setting, the accumulation of errors from the first order Taylor approximation along a long path leads to worse sampler efficiency despite better exploration.
Moreover, using a longer path length leads to discovering variants with a higher number of mutations, whose activity could be inaccurately characterized by the supervised experts.

\subsubsection{Run time comparison}
We compare wall clock run times for PPDE and the baseline samplers under the Potts (\Figref{fig:potts-compute}) and ESM2 (\Figref{fig:esm2-compute}) experts using the PABP protein, 1K sampler steps, and a population size of 16 variants.
Samplers are run on a single NVIDIA Tesla cloud GPU; see appendix for complete hardware details.
\texttt{CMA-ES} is significantly slower than the other samplers.
When using the Potts expert, \texttt{PPDE}'s run time is relatively comparable to \texttt{simulated annealing} and \texttt{MALA-\emph{approx}}. 
The cost of computing the gradient of ESM2 is much higher than for the Potts model, so in this setting \texttt{simulated annealing} (which does not use gradients) has lower run time than \texttt{MALA-\emph{approx}} and \texttt{PPDE}.  

\section{Limitations}
\label{sec:limits}
In this section, we discuss limitations of the proposed sampling method related to computational costs and handling insertion/deletion mutations.

Each step of the PPDE sampler requires two backwards passes---one for computing the forward proposal and one for the reverse proposal. 
When PPDE is used with a large protein language model whose backwards pass has a large memory footprint, this becomes a bottleneck if sufficient computing resources are unavailable.
Fortunately, MCMC samplers are embarrassingly parallel.
For example, with a budget of $K$ available GPUs, we can evolve a single protein per GPU in parallel.

The wall-clock time incurred by running PPDE depends on the ability to rapidly evaluate and differentiate through the product of experts distribution.
For certain classes of unsupervised experts, this is difficult.
For example,  evaluating the log probability of a protein with a variational autoencoder (e.g., DeepSequence~\citep{Riesselman2018}) requires computing a large-sample Monte Carlo estimate of the evidence lower bound, which is highly computationally expensive.

PPDE also does not currently support inserting and deleting amino acids (indels).
We necessarily assume the product of experts is locally smooth in a neighborhood around the current protein to compute our proposal distribution for MCMC.
This assumption would need to still hold despite proposing length changes to the sequence.
While we do not believe lack of support for indels is a fundamental limitation of the sampler, supporting indel mutations appears to be nontrivial.

\section{Conclusions}

In this study, we have shown how to flexibly combine unsupervised models of evolutionary density, including protein language models, and supervised models of protein function and how to efficiently sample from the resulting distribution to discover proteins that maximize a desired function while avoiding poor local optima.
This strategy leverages the vast amounts of unlabeled data that are available for unsupervised (or self-supervised) pre-training to improve designed sequences, even when relatively few labelled data are available for training the fitness function. 
Our empirical results suggest our framework offers a practical and effective new paradigm for machine-learning-based directed evolution.
Although our framework can potentially be applied to a broader range of biological sequence design tasks, we focused solely on proteins due to the variety of available pre-trained models.

Based on our findings, we recommend constructing the product of experts distribution with unsupervised experts that have been pre-trained on \emph{aligned} protein sequences (e.g., MSAs) when available.
We found that these models more strongly constrain search to the evolutionarily plausible mutations than unsupervised models pre-trained on \emph{unaligned} protein sequences.
However, \emph{combining} unsupervised models of both types produced designs superior to using one or the other alone.

One important capability that we did not demonstrate is the incorporation of additional hard constraints, such as a maximum allowable number of mutations per variant or the preservation of particular regions of the WT sequence. 
In our framework, we can easily accomplish this by forcing the mutation proposal probabilities at the sequence positions in question to zero, which can be done by setting their logits to negative infinity in the softmax in Algorithm~\ref{alg:pppo}.
Future work may also extend this framework to larger problems in biological design.
For instance, the simultaneous engineering of several sequences in multimeric enzyme complexes, or incorporating substrate structure in evaluating the likelihood of enzyme-substrate complexes.

\section*{Acknowledgments}
This work was authored by the National Renewable Energy Laboratory, operated by Alliance for Sustainable Energy, LLC, for the U.S. Department of Energy (DOE) under Contract No. DE-AC36-08GO28308. Funding provided by the U.S. Department of Energy Bioenergy Technologies Office and the Laboratory Directed Research and Development (LDRD) Program at NREL. The views expressed in the article do not necessarily represent the views of the DOE or the U.S. Government. The U.S. Government retains and the publisher, by accepting the article for publication, acknowledges that the U.S. Government retains a nonexclusive, paid-up, irrevocable, worldwide license to publish or reproduce the published form of this work, or allow others to do so, for U.S. Government purposes.
This research used resources of the Oak Ridge Leadership Computing Facility at the Oak Ridge National Laboratory, which is supported by the Office of Science of the U.S. Department of Energy under Contract No. DE-AC05-00OR22725.

\section*{Data availability statement}
The data and code that support the findings of this study are openly available at the following URL: \url{https://github.com/pemami4911/ppde}.

\bibliography{main}

\newpage
\appendix

\section{Deriving the Gradient-based Approximate Proposal}
\label{sec:app:derivation}

We provide a short derivation that illustrates how to obtain a gradient-based approximate locally-balanced informed proposal (\Eqref{eq:gwg}) from the locally-balanced informed proposal of \Eqref{eq:zanella}. The goal is to avoid evaluating $f(x')$ for all $x' \in \mathcal{N}(x)$ when computing the proposal distribution $q(x'|x) \propto \exp(f(x') - f(x))^{\frac{1}{2}}\textbf{1}(x' \in \mathcal{N}(x))$. The basic idea from~\citet{Grathwohl2021} is to assume that $f$ is a continuously differentiable function, which allows us to approximate $f(x') - f(x)$ with a first-order Taylor series. 

In detail, the first-order Taylor series of the function $f(x)$ at the point $a$ is
\begin{equation}
    f(x) \approx f(a) + \nabla_{a}f(a)^{\intercal}(x - a) .
\end{equation}
By subtracting $f(a)$ from both sides, we get
\begin{equation}
    f(x) - f(a) \approx \nabla_a f(a)^{\intercal}(x - a).
\end{equation}
To arrive at the gradient-based proposal, suppose $a$ is the current state of our MCMC sampler and let $x$ be any point in the neighborhood of $a$, $x \in \mathcal{N}(a)$. 
When we plug in our first-order approximation,
\begin{equation}
    \Tilde{q}(x|a) \propto \exp( \nabla_a f(a)^{\intercal}(x - a) )^{\frac{1}{2}}\textbf{1}(x \in \mathcal{N}(a)).
\end{equation}

\section{Proof of Corollary 1}
\label{sec:app:proof}
The basic idea of the proof is to use the triangle inequality to obtain a bound on the approximation error of a \emph{sum} of experts which assumes that each expert has sufficiently smooth gradients.

\textbf{Definition} A function $f : \mathbb{R}^N \rightarrow \mathbb{R}$ has \emph{$K$-Lipschitz continuous gradient} when
\begin{equation}
    \lVert \nabla_{x'} f(x') - \nabla_{x} f(x) \rVert \leq L \lVert x' - x \rVert
\end{equation}
for all $x,x' \in \mathbb{R}^N$.

For convenience, we reproduce a pertinent result here from \citet{Nesterov} (Lemma 1.2.3). 

\textbf{Lemma 1.2.3}~\citet{Nesterov} \emph{If $f : \mathbb{R}^N \rightarrow \mathbb{R}$ has an $L$-Lipschitz gradient, then for any $x,x' \in \mathbb{R}^N$ we have}:
\begin{equation}
    \label{eq:nesterov-lemma}
    | f(x') - f(x) - \langle \nabla_x f(x), x' - x \rangle | \leq \frac{L}{2} \lVert x' - x \rVert ^2.
\end{equation}

\begin{lemma}
For functions $f$ and $g$ with $K_f$-Lipschitz and  $K_g$-Lipschitz gradients respectively, the sum-composition $h = f + g$ has $(K_f + K_g)$-Lipschitz gradient.
\label{eq:composition}
\end{lemma}

\emph{Proof}: By the triangle inequality:
\begin{align}
\lVert \nabla_{x'} h(x') - \nabla_x h(x) \rVert &= \lVert \nabla_{x'} (f(x') + g(x'))- \nabla_x (f(x) + g(x)) \rVert \nonumber  \\
&=
\lVert \nabla_{x'} f(x') + \nabla_{x'} g(x') - \nabla_x f(x) - \nabla_x g(x) \rVert \nonumber \\
&\leq \lVert \nabla_{x'} f(x') - \nabla_x f(x) \rVert + \lVert \nabla_{x'} g(x') - \nabla_x g(x) \rVert \nonumber\\
&\leq (K_f + K_g) \lVert x' - x \rVert.
\end{align}

\begin{lemma}
Suppose $f_i$, $i=1,\dots,M$ are functions with $K_i$-Lipschitz gradient. Then for any $x,x' \in \mathbb{R}^N$ we have
\begin{equation}
    \Biggl| \sum_{i=1}^M \Big( f_i(x') - f_i(x)\Big) - \Big \langle \sum_{i=1}^M \nabla_x f_i(x), x' - x \Big \rangle \Biggr| \leq \frac{\sum_{i=1}^M K_i}{2} \lVert x' - x \rVert ^2.
\end{equation}
\label{eq:geometry}
\end{lemma}

\emph{Proof:} Let $g = \sum_{i=1}^M f_i$ where each $f_i$, $i=1,\dots,M$ has $K_i$-Lipschitz gradient. By Lemma 1 we can see that $g$ has a $\sum_{i=1}^M K_i$-Lipschitz gradient. Then Lemma 2 follows immediately by applying Lemma 1.2.3 from \citet{Nesterov} to $g$.

The proof of Corollary 1 proceeds by bounding the approximation error between two consecutive states $x^{r-1}$ and $x' \in \mathcal{N}(x^{r-1})$ in a path of length $R \sim \text{Unif}(1,U)$. For simplicity we assume $\mathcal{N}(x)$ is the 1-Hamming ball, i.e., $\lVert x^r - x^{r-1}  \rVert ^2 = 1$.

For $g = \sum_{i=1}^M f_i$ which has $K = \sum_{i=1}^M K_i$-Lipschitz gradient, Lemma 2 gives us that
\begin{align*}
    -\frac{K}{2}  \leq g(x') - g(x^{r-1}) - \langle \nabla g(x^{r-1}), x' - x^{r-1} \rangle \leq \frac{K}{2} .
\end{align*}
Then an upper bound for $g(x') - g(x^{r-1})$ is
\begin{align*}
    g(x') - g(x^{r-1}) &\leq \langle \nabla g(x^{r-1}), x' - x^{r-1} \rangle + \frac{K}{2} \\
    &= \langle \nabla g(x^0), x' - x^{r-1} \rangle + \langle \nabla g(x^{r-1}) - \nabla g(x^0), x' - x^{r-1} \rangle + \frac{K}{2} \\
    &\leq  \langle \nabla g(x^0), x' - x^{r-1} \rangle + K r + \frac{K}{2} \\
    &= \langle \nabla g(x^0), x' - x^{r-1} \rangle + K ( r - \frac{1}{2}).\\
\end{align*}
Following similar steps, we also have
\begin{align*}
    g(x') - g(x^{r-1}) \geq \langle \nabla g(x^0), x' - x^{r-1} \rangle + K  ( r + \frac{1}{2}).
\end{align*}

The remainder of the proof for Corollary 1 exactly follows Equations 64-72 in the proof of Theorem 3 in \citet{sun2022path} with $g(x)$ which has $K$-Lipschitz gradient.

\section{Experiment details} 

\subsection{Neural architectures}

\textbf{MNIST denoising autoencoder: } 
The encoder and decoder neural networks are constructed out of residual blocks, where a single block has two 3x3 2D Conv layers, each followed by a BatchNorm layer and Swish activation. The first Conv layer in the block uses a stride of 2 and a 1x1 shortcut Conv layer is added to its output.
The encoder consists of one 3x3 2D Conv layer, followed by two blocks, then another 3x3 2D Conv layer, followed by a fully connected layer that projects the flattened output features into a 16-dimensional latent space.
The encoder and decoder are symmetric, and the decoder is implemented with transposed Conv layers and the per-pixel output is interpreted as the logits of a Bernoulli distribution.
All Conv layers use 64 channels.
To train the DAE, we corrupt the input $28 \times 28$ binary MNIST image by randomly flipping $p$\% of the pixels, where $p \sim $ Unif(0,15), and minimize the reconstruction error. 
We use the AdamW optimizer with a learning rate of 1e-4 to train the model for 10K steps using a mini-batch size of 100.

\noindent \textbf{MNIST energy-based model: }This model follows the residual EBM architecture from~\citet{Grathwohl2021} closely.
The model consists of one 3x3 2D Conv layer, followed by 2 residual blocks  (the same blocks used by the DAE), and then an additional 6 residual blocks with all Conv layers using a stride of 1.
The output features are flattened with global average pooling and then mapped to a scalar with a fully connected layer.
See~\citet{Grathwohl2021} for details about the contrastive divergence training algorithm and training hyperparameters. 

\noindent \textbf{Protein ConvNet: }We use the ConvNet baseline from the FLIP benchmark~\citep{dallago2021flip} to regress activity. This model takes in a mini-batch of one-hot encoded protein sequences of shape $B \times L \times V$, applies a 1D Conv with kernel size 5 and ReLU activation to project $V$ to dimension $L$, followed by a fully connected layer to expand the feature dimension to $2L$.
Then, max pooling is used to reduce the the length of the sequence to 1, and a linear layer projects the feature dimension from $2L$ to 1.
The model is trained with AdamW optimizer with a learning rate of 1e-3 using a mini-batch size of 256.

\subsection{Unsupervised expert score functions}
We provide details for computing $\sum_i f_i(x)$ with each type of unsupervised expert.

\noindent \textbf{MNIST denoising autoencoder: }For a given image $x$, we define $f(x)$ to be the sum of the binary cross entropy between the input pixel and the predicted pixel logits over all pixel locations.

\noindent \textbf{MNIST energy-based model: }We use the scalar output as $f(x)$, since it can be interpreted as an unnormalized log probability for the input. This only requires a single forward pass to compute.

\noindent \textbf{Protein EVmutation Potts: }We use the difference in the Potts Hamiltonian between the protein variant $x$ and the wild type such that $f(x) = H(x) - H^{WT}(x)$. 
The Hamiltonian for a one-hot protein $x$ is defined as 
\begin{equation}
    H(x) = \sum_{i=1}^L h_i^T x_i + \sum_{i,j=1}^L x_i^T J_{ij} x_j \nonumber
\end{equation}
where $J \in \mathbb{R}^{L \times L \times 20 \times 20}$ and $h \in \mathbb{R}^{L \times 20}$ are the Potts parameters.

\noindent \textbf{Protein ESM2: }We use the difference in the sum of the per-amino-acid log probabilities between the protein variant $x$ and the wild type for $f(x)$. 
For computational efficiency, we compute the score with a single forward pass and therefore do not use any masking of input sequence positions.
In detail, for a one-hot protein $x$ let $\phi(x)$ be the $L \times V$ matrix of logits computed by a single unmasked forward pass of ESM2.
Then the score is
\begin{equation}
    f(x) = \sum_{i=1}^L \sum_{j=1}^V x_{ij} \texttt{log\_softmax}(\phi(x)_i)_{j} - \sum_{i=1}^L \sum_{j=1}^V x^{WT}_{ij} \texttt{log\_softmax}(\phi(x^{WT})_i)_{j}. \nonumber
\end{equation}

\subsection{Hardware details}

We ran our experiments on the Summit supercomputer on a single NVIDIA V100 GPU.
When running with ESM2, we speed up the sampler by parallelizing the population of 128 proteins across multiple GPUs.
For example, we ran each of the 128 MCMC trajectories on a separate GPU in parallel when we used the 650M parameter ESM2 model.
Wall clock run time measurements were taken after-the-fact using a single NVIDIA Tesla T4 GPU available via the Google Colab free tier.

\subsection{Aligning the amino acid vocabulary across models}

Not all pre-trained protein sequence models order the amino acids in their vocabulary in the same way. 
Also, some models, such as masked language models, have extra tokens such as \texttt{<mask>} beyond the standard 20 amino acids. 
To enable combining a diverse array of models, we first decide on a canonical ordering for the standard amino acids.
For any pre-trained model that uses a different ordering than the canonical one, we pre-compute a permutation matrix that, when applied to the columns of the one-hot encoded protein $x$, will put the amino acids in the order which the model expects.
We also pad the permuted $x$ with extra columns of zeros if the pre-trained model has a vocabulary larger than 20.

\begin{landscape}
\begin{table}[t]
\small
\centering
\begin{tabular}{@{}lccccccccc@{}}
\toprule
   &
   \multicolumn{3}{c}{\begin{tabular}[c]{@{}c@{}}Log fitness\small{$\uparrow$}\\(Augmented EVmutation)\end{tabular}} &
   \multicolumn{3}{c}{\begin{tabular}[c]{@{}c@{}}Evolutionary density\small{$\uparrow$}\\(MSA Transformer)\end{tabular}} & \multicolumn{3}{c}{\begin{tabular}[c]{@{}c@{}}Exploration \\(mean\tiny{$\pm$std }\normalsize  \# muts)\end{tabular}} \\ \midrule
    Potts expert & PABP & UBE4B & GFP & PABP & UBE4B & GFP & PABP & UBE4B & GFP  \\ \midrule 
PPDE &    0.27\tiny{(0.86)}  &    0.39\tiny{(1.18)}   &  -0.04\tiny{(0.24)}   &   6.92\tiny{(13.43)}   &  -4.79\tiny{(-2.14)}     &   -5.98\tiny{(-0.76)}  & 3.5\tiny{$\pm$ 0.9} & 2.7\tiny{$\pm$ 0.6} & 2.0\tiny{$\pm$ 0.3} \\
Random search & 0.09\tiny{(0.82)} & -0.19\tiny{(0.34)} & -0.04\tiny{(0.04)} & 4.26\tiny{(6.87)} & -1.09\tiny{(2.46)} & -0.11\tiny{(-0.11)} & 1.3\tiny{$\pm$ 0.5} & 1.1\tiny{$\pm$ 0.3} & 1.0\tiny{$\pm$ 0.2}\\
Sim. annealing   &  0.09\tiny{(0.44)}    &  -0.19\tiny{(0.28)}   &  -0.04\tiny{(0.10)}  &  3.55\tiny{(8.63)}    &  -0.94\tiny{(2.70)}  & -5.89\tiny{(-0.99)}  &  1.3\tiny{$\pm$ 0.5} & 1.0\tiny{$\pm$ 0.2} & 1.0 \tiny{$\pm 0.1$}   \\
MALA\emph{-approx}   &   0.09\tiny{(0.56})   &  -0.19\tiny{(0.58)}      &  -0.04\tiny{(0.10)}    & 2.25\tiny{(5.14)}     &  -0.89\tiny{(1.54)}   &   -6.74\tiny{(-1.88)}   &   1.3\tiny{$\pm$ 0.5} & 1.03\tiny{$\pm$ 0.2} & 1.03\tiny{$\pm$ 0.2}\\
CMA-ES               &  1.37\tiny{(1.37)}    &  2.54\tiny{(2.54)}      &   -2.50\tiny{(-0.15)}  & 3.47\tiny{(3.47)}     &   -94.76\tiny{(0.0)}    & -62.43\tiny{(0.0)}    &   17.0\tiny{$\pm$ 0} & 15.5\tiny{$\pm$ 6.2} & 10.2\tiny{$\pm$ 9.4}  \\
\midrule
Unsupervised expert & & & & & & & & &\\
\midrule
Potts only & 0.70\tiny{(1.47)} & 0.12\tiny{(0.99)} & -0.18\tiny{(0.21)} & 9.17\tiny{(18.54)} & -4.24\tiny{(-2.68)} & -1.88\tiny{(-1.59)} & 4.7 \tiny{$\pm$ 1.2} & 2.6 \tiny{$\pm$ 0.5} & 1.2 \tiny{$\pm$0.4} \\
Supervised only         &  0.14\tiny{(0.44)}    &   1.66\tiny{(5.26)}    &  -0.23\tiny{(0.14)}   &   -2.56\tiny{(0.48)}   &  -6.83\tiny{(-6.29)}     & -9.28\tiny{(-2.13)}    &   1.9 \tiny{$\pm$ 0.8} & 1.3 \tiny{$\pm$ 0.7} & 1.7\tiny{$\pm$ 0.8}   \\
ESM2          &  0.14\tiny{(0.63)}    &  1.66\tiny{(5.56)}     & -5.55\tiny{(0.16)}     &   -2.38\tiny{(5.56)}   &   -6.58\tiny{(-3.83)}    &   -126.82\tiny{(5.90)}  &   2.9\tiny{$\pm$ 1.3} & 2.2\tiny{$\pm$ 2.2} & 14.9\tiny{$\pm$ 12.6} \\
Potts+ESM2    & 0.44\tiny{(1.48)} & 1.30\tiny{(3.33)} & -0.04\tiny{(0.33)}   &  9.12\tiny{(19.34)}    &  -13.6\tiny{(1.98)}     &  -7.17\tiny{(8.11)}     &   5.3\tiny{$\pm$1.8}   & 4.3\tiny{$\pm$0.7}  & 2.1\tiny{$\pm$0.3}     \\
\bottomrule
\end{tabular}
\caption{\textbf{50\textsuperscript{th}\tiny{(100\textsuperscript{th})} \normalsize{percentile scores}}. Population size is 128. Across all three proteins, \texttt{PPDE (Potts)} and \texttt{PPDE (Potts+ESM2)} in particular discover variants with higher predicted fitness and average \# of mutations than \texttt{Random search}, \texttt{Simulated annealing}, and \texttt{MALA-\emph{approx}}. 
Out of the samplers using the Potts expert, 
\texttt{PPDE (Potts)} achieves the highest evolutionary density scores on PABP.
We suspect that the slightly lower 50\textsuperscript{th} percentile evolutionary density scores compared to the baselines (except \texttt{CMA-ES}) on the more challenging UBE4B and GFP proteins are in part due to PPDE discovering variants with more average mutations ($2-4+$ mutations vs. $\sim 1$ mutation per variant) and higher fitness.
The 100\textsuperscript{th} percentile performance of \texttt{PPDE (Potts+ESM2)} shows impressively high log fitness \emph{and} evolutionary density, however.
\texttt{CMA-ES} sampler had inconsistent performance across the three proteins---it finds variants with high numbers of mutations ($10-17$) but low evolutionary density scores (3.47 on PABP, -94.76 on UBE4B and -62.43 on GFP compared to 6.92, -4.79, and -5.98 for \texttt{PPDE (Potts)}).
It finds just one protein variant for PABP that is 17 mutations from WT with an evolutionary density score of 3.47, and it seems to have significant difficulty with the larger proteins UBE4B and GFP; e.g., on GFP the average 50\textsuperscript{th} percentile log fitness is -2.50 compared to -0.04 for PPDE. We conclude that \texttt{CMA-ES} can be recommended for use only when a single protein variant is desired and the length of the protein is relatively small (e.g., $\leq$ 95 residues). \label{tab:main}}
\end{table}

\end{landscape}

\end{document}